\documentclass[10pt,twocolumn,letterpaper]{article}

\usepackage[pagenumbers]{cvpr} 

\usepackage[accsupp]{axessibility}
\usepackage{graphicx}
\usepackage{amsmath}
\usepackage{amssymb}
\usepackage{booktabs}
\usepackage{upgreek}
\usepackage{enumerate}
\usepackage{stmaryrd}
\usepackage{bbold}
\usepackage{url}
\usepackage{soul}
\usepackage[table]{xcolor}
\usepackage{pifont}
\usepackage{multirow}
\usepackage{comment}

\soulregister\cite7
\soulregister\ref7

\newcolumntype{I}{!{\vrule width 1.2pt}}
\newlength\savedwidth
\newcommand\whline{\noalign{\global\savedwidth\arrayrulewidth
		\global\arrayrulewidth 1.25pt}%
	\hline
	\noalign{\global\arrayrulewidth\savedwidth}}

\usepackage[pagebackref,breaklinks,colorlinks]{hyperref}

\usepackage[capitalize]{cleveref}
\crefname{section}{Sec.}{Secs.}
\Crefname{section}{Section}{Sections}
\Crefname{table}{Table}{Tables}
\crefname{table}{Tab.}{Tabs.}

\def\cvprPaperID{5385} 
\def\confName{CVPR}
\def\confYear{2022}

\makeatletter

\newcommand{\Rmnum}[1]{\expandafter\@slowromancap\romannumeral #1@}
\makeatother

\begin{document}

\title{DR.VIC: Decomposition and Reasoning for Video Individual Counting}
\author{Tao Han$^{1\dag}$, \quad Lei Bai$^{2\dag}$, \quad Junyu Gao$^{1}$, \quad Qi Wang$^{1\ast}$, \quad Wanli Ouyang$^{2}$ \\
	$^1$   Northwestern Polytechnical University, Xi’an
	710072, P.R. China\\ $^2$ The University of Sydney, SenseTime Computer Vision Group, Australia\\
	{\tt\small hantao10200@mail.nwpu.edu.cn, \{baisanshi, gjy3035, crabwq\}@gmail.com,} \\ 
	{\tt\small wanli.ouyang@sydney.edu.au }
}
\maketitle

\begin{abstract}
{Pedestrian counting is a fundamental tool for understanding pedestrian patterns and crowd flow analysis.
Existing works (e.g., image-level pedestrian counting, cross-line crowd counting \etal) either only focus on the image-level counting or are constrained to the manual annotation of lines.
In this work, we propose to conduct the pedestrian counting from a new perspective - Video Individual Counting (VIC), which counts the total number of individual pedestrians in the given video (\textbf{a person is only counted once}).
Instead of relying on the Multiple Object Tracking (MOT) techniques, we propose to solve the problem by decomposing all pedestrians into the initial pedestrians who existed in the first frame and the new pedestrians with separate identities in each following frame. Then, an end-to-end \underline{D}ecomposition and \underline{R}easoning \underline{Net}work (\textbf{DRNet}) is designed to predict the initial pedestrian count with the density estimation method and reason the new pedestrian's count of each frame with the differentiable optimal transport. 
Extensive experiments are conducted on two datasets with congested pedestrians and diverse scenes, demonstrating the effectiveness of our method over baselines with great superiority in counting the individual pedestrians. Code: \url{https://github.com/taohan10200/DRNet}.}
\end{abstract}

\renewcommand{\thefootnote}{\fnsymbol{footnote}}
\footnotetext[2]{\quad Equal contribution.}
\footnotetext[1]{\quad Corresponding author.}
\section{Introduction} \label{sec:intro}
The world population has witnessed rapid growth, along with the accelerated urbanization. It is expected that around 70\% of the world's population will live in cities~\cite{un2019world,chaudhary2021video}, which brings significant challenges in the city management, such as transport management, public space design, evacuation planning, and public safety. To tackle these challenges, accurately obtaining the number of pedestrians of any region in a period of time, e.g., the number of people passed the intersection in the past 10 minutes, is a basic problem.

\begin{figure}[t]
	\centering
	\includegraphics[width=0.45\textwidth]{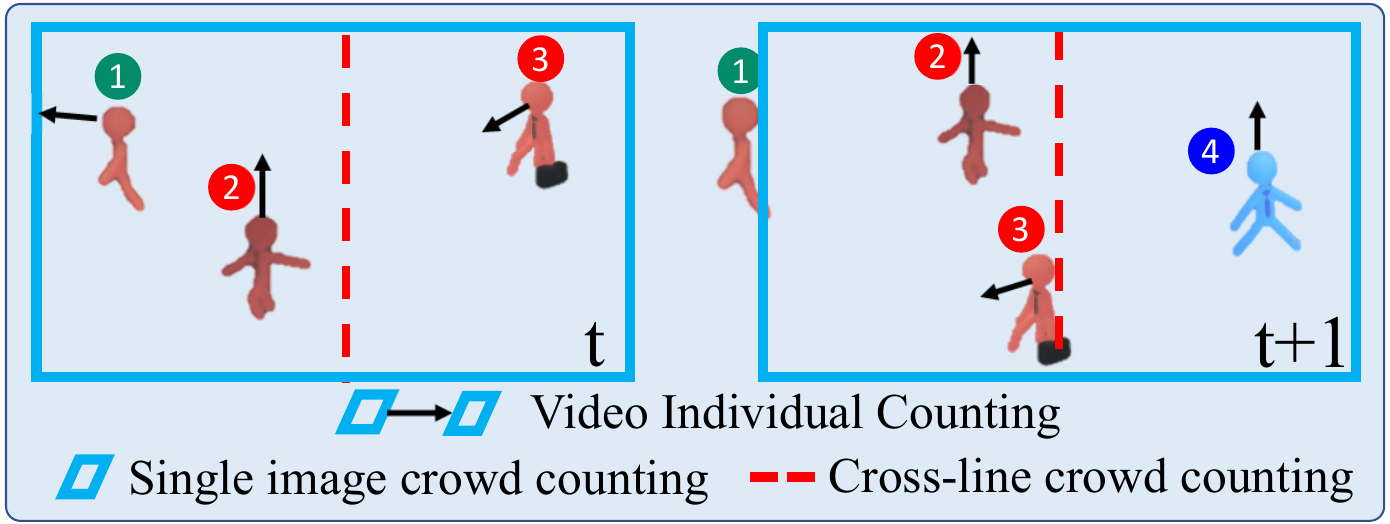}
	\caption{Illustration of different crowd counting paradigms. Video Individual Counting count all pedestrians appearing in the video and each person is only counted once, e.g. count \ding{172}\ding{173}\ding{174}\ding{175} to get counting result 4. Single Image Crowd Counting targets on the image-level. Directly applying it to video individual counting will cause over-count, e.g., count pedestrians \ding{172}\ding{173}\ding{174} at frame $t$ and \ding{173}\ding{174}\ding{175} at frame $t+1$ getting result 6. Cross-line Crowd Counting targets on the video level but only focuses on the pedestrians passed the red virtual line, e.g., count \ding{174} to get result 1.}  
	\vspace{-0.5cm}
	\label{Fig:intro}
\end{figure}

Automatically estimating the pedestrian number from images/videos is a practical solution and attracts attentions of researchers from different perspectives. Specifically, Single Image Crowd Counting (SICC) \cite{chan2008privacy,li2018csrnet,wan2019residual, ma2019bayesian,gao2020feature} estimates the crowd number in the image level, which can reflect the degree of crowdedness at a certain time point. Video Crowd Counting (VCC) \cite{fang2019locality,wu2020fast,xiong2017spatiotemporal} further enhances SICC by exploring the information from the historical frames to achieve a more accurate and robust counting in the target frame. 
Different from SICC and VCC, the cross-line crowd counting techniques \cite{cong2009flow,ma2015counting,zhao2016crossing,zheng2018cross} focuses on estimating the pedestrians in a period of time from videos. By manually setting a proper virtual line (e.g., the red line as illustrated in Fig.~\ref{Fig:intro}), cross-line crowd counting discovers the pedestrians passed this line, which could reflect the crowdness and total crowd number in the video. 

Different from the settings of the works discussed above, this work targets a similar but different task --Video Individual Counting (VIC), which counts the total number of pedestrians in the given video with separate identities. As illustrated in \cref{Fig:intro}, people come from all directions into the camera view should be counted through time and each person should only be counted once.
The output of VIC is the common concept regarding the total pedestrian number in a scene during a period of time (i.e., video length), which accurately measures the crowdness and popularity of an area, and thus is valuable for numerous applications, e.g., count the people attending an event, or measure the total pedestrians of an intersection.

While sharing similarities, Video Individual Counting is challenging and cannot be directly solved by existing image-level pedestrian counting and cross-line counting methods. Image level pedestrian counting methods will inevitably count the same person multiple times in adjacent frames and hugely over-estimates the pedestrian number if directly summing the number of crowd at each frame. Cross-line counting methods only count the pedestrians passing the line and thus will miss the crowd staying in the video or disjointing with the line. Besides, it requires manual annotation of the line for different camera settings, which is not in accordance with the aim of computer vision researchers in making crowd counting automated, and is also time consuming when considering a single city like London might have 500,000+ CCTV cameras.

The potential existing solution for VIC is the Multi-Object Tracking (MOT) methods \cite{zhang2021fairmot,sundararaman2021tracking}, which is a general technique to identify and track all objects in the video and has been explored for specific area (e.g., counting for bus entrance~\cite{sun2019benchmark}). The number and states of tracks in MOT can be employed to not only reveal the total number of crowd in the video, but also the inflows and outflows (the number of people getting into and out of the scene, respectively) of a period. However, since MOT is designed for tracking instead of pedestrian counting, the accuracy and efficiency of using MOT for this task would be inferior for two reasons. For accuracy, the object association in MOT depends on the detection results of multiple frames and extremely suffers from the ID switch, which heavily influence the counting performance by over-counting. For efficiency, most MOT operates on each frame, which is time-consuming and not necessary for the crowd counting task.

We propose a new paradigm for Video Individual Counting without relying on the MOT.
Instead of associating all pedestrians in the whole video as MOT, we only associate each pair of frames to identify the inflow (i.e., new pedestrians) of each time slot.
Specifically, we decompose all pedestrians into the initial pedestrians existed in the first frame and the new pedestrians with separate identities in the following frames.
The rational behind this idea is the observation that the crowd normally stay in or pass through a region (e.g., camera view region). Only in few cases, people may pop in and out of the camera view, which is neglectable compared with the counting error. Take CroHD \cite{sundararaman2021tracking} dataset as an example, only 17 pedestrians leave and re-enter the camera view region after an interval of 75 frames, while the total pedestrians number in the videos is 5230.

Based on this idea, an efficient and end-to-end video individual counting framework named \underline{D}ecomposition and \underline{R}easoning \underline{Net}work (DRNet) is proposed.
DRNet first samples frames from the video with a time window. Then each pair of frames formed by the adjacent frames are separately fed into neural networks for generating two CNN feature maps, based on which two density maps reflecting the head locations at each frame can be predicted separately.
In the next step, two sets of features containing the descriptors of each located head are sampled from the CNN feature maps and then used by a pedestrian inflow reasoning module implemented with differentiable Optimal Transport, from which the inflow (pedestrians joining the latter frame) and outflow (pedestrians leaving the former frame) of the input frame pair can be predicted.
Finally, the total pedestrian count in the whole video can be obtained by integrating the crowd count of the first frame and all pedestrian inflows in the following sampled frame pairs.

Our core contributions are summarized as follows:
\begin{itemize}
\setlength{\itemsep}{0pt}
\setlength{\parsep}{0pt}
\setlength{\parskip}{0pt}
  \item We propose to decompose video individual pedestrians into the initial pedestrians and the new pedestrians at following frames, which avoids the complicated and error-prone video-level association in MOT and opens a new direction for pedestrian counting.
  \item We propose an efficient and end-to-end framework to achieve video individual counting, directly obtaining the new pedestrians of a frame by reasoning it with the preceding frame using differentiable optimal transport.
  \item Extensive experiments are conducted on two datasets covering congested crowds and diverse scenes. The experimental results demonstrate the effectiveness of the proposed methods over strong baselines.
\end{itemize}

\section{Related Work}
\subsection{Image-level crowd counting}
Image-level crowd counting refers to counting the number of people in a given static crowd image.
In recent years, most state-of-the-art SICC methods concentrate on density map estimation, which integrates the density map as a count value. The CNN-based methods \cite{zhang2016single,liu2019crowd,li2018csrnet,liu2021cross,bai2020adaptive,9337191, zhang2021cross} show the powerful ability in feature extracting and representing than the models based on hand-crafted features \cite{idrees2013multi,liu2015bayesian}. 
Apart from the density-map supervision, some methods \cite{ma2019bayesian,wang2020distribution,liu2019point,dong2020scale} directly exploit point-level labels to supervise counting models. Because density maps can only give a coarse count, the location distribution of people is still not available. Therefore, latest researches \cite{idrees2018composition, liu2019recurrent, gao2020nwpu,abousamra2020localization,wan2021generalized,song2021rethinking,sam2020locate,gao2020learning} target to localization for counting.

\textbf{Image-Level counting from multiple frames.} There are also some video-based crowd counting methods \cite{6520940,fang2019locality,wu2020fast,xiong2017spatiotemporal,liu2020estimating}, which exploit the temporal information to enhance the counting in the target frame. 

Different to these works, our work focuses on individual pedestrian counting in the video level, which predicts the total number of the dynamic people over video frames. As discussed before, the targeted task of video individual counting cannot be solved by directly using image-level counting methods. However, since image-level counting methods can be used as a basic component in our design, the progress in image-level crowd counting can benefit our model for the sub-tasks of initial counting in the first frame and the head localization.

\subsection{Video-level crowd counting}

\subsubsection{Tracking in crowd}
\label{sec2:crowd_localizaiton}
Tracking in crowd \cite{luo2020multiple,8283783} means to extract the temporal information of the crowd in a continuous sequence of images in a video.
Considering the group motion behavior is consist of individual behaviour, Kratz and Nishino \cite{kratz2009anomaly} propose a Bayesian framework that uses a space-time model for tracking individuals in a crowd. Bera \etal \cite{bera2014adapt} propose a real-time algorithm, AdaPT, to calculate individual trajectory in a crowd scene. SSIC can be integrated to crowd tracking as well, Ren \etal \cite{ren2020tracking} propose a tracking-by-counting method, which jointly model detection, counting, and tracking for capturing complementary information.
Since tracking explicitly distinguish the identity of each person in the video, it can also be used for real-time people counting. Sun \etal \cite{sun2019benchmark} propose a RGB-D dataset that collected from the bus entrance door in surveillance view and utilize tracking to identify and count the entering and exiting people. Recently, Sundararaman \cite{sundararaman2021tracking} \etal propose a congested Heads Dataset (CroHD), a head detector with a Particle Filter, and a color histogram based re-identification module to track multiple people in crowded scene. 

While our method also calculates cross-frame association, it does not rely on detection or video-level association. We only utilize cross-frame association to get the inflows of each time interval and integrate them together with the counting in the first frame to get the total individual number in the whole video. In this way, our design is more robust to detection and tracking errors. 

\subsubsection{Cross-line crowd counting}
Cross-line crowd counting is a constrained direction of video pedestrian counting, which aims to count the number of pedestrian across a detection line or inside a specified region. Early works  \cite{albiol2001real,bescos2003dct,barandiaran2008real} utilize multiple lines to count the entering or existing people. These methods, however, need to perform counting independently for each line belonging to the counting zone, which is inefficient. Besides, it does not count the people who stay in the scene but do not cross the line. More importantly, it does not allow to sample the frame for reducing temporal redundancy with long interval. Cong \etal \cite{cong2009flow} regard pedestrians across the line as fluid flow and devise a algorithm to estimate the flow velocity field. The final count is obtained by integrating the pixels in Line of Interest (LOI) at each frame. To tackle some drawbacks in blob-centric method \cite{cong2009flow}, Ma \etal \cite{ma2015counting} propose an integer programming approach to estimate the instantaneous counts on the LOI, which cuts the frames in a video to a set of temporal slices and then counts the people in these slice images with SICC methods. To further eliminate the limitation of the temporal slice, Zhao \etal \cite{zhao2016crossing} propose to directly estimate the crowd counts with pair of images, which resolves the LOI crowd counting by estimating the pixel-level crowd density map and crowd velocity map. The following work \cite{zheng2018cross} further improves the \cite{zhao2016crossing} to obtain a fine-grained estimation of local crowd densities.

In general, cross-line crowd counting is limited in real application as it highly depends on the virtual line, which is hard to set for numerous videos and capture pedestrians entering and existing with random directions (e.g., squares). In contrast, our method can handle multi-direction pedestrians, and thus is applicable to more general scenes.
\label{sec:crowd_detection}

\section{Problem Formulation} \label{Sec:Prolem}
Given a video clip $\mathcal{I} = \{\mathbf{I}_0, \mathbf{I}_1, ..., \mathbf{I}_{T-1}\}$ of length $T$ for a scene (e.g., intersection, square, exhibition), where the $t$-th frame $\mathbf{I}_t$ contains $N(t)$ subjects, each subject normally appears in many consecutive frames because of the relatively high sensing frequency of cameras (e.g., 25 frames per second). Our target is to count the total number of people $N(0:T-1)$ shown in this video with distinct identities.

Instead of relying on the MOT techniques to directly obtain the crowd count with the track number at $T$, we propose a new solution for video individual counting. Specifically, we formulate the video individual counting problem as two sub-problems: 1) inferring crowd count  $N(0)$ at the start time point, and 2) identifying the number of new identities entering the scene (inflow) at each following frame $N_{in}(t)$, which requires associating the subjects in frame $t-1$ and $t$. By solving these two sub-problems, the overall video pedestrian count can be easily obtained via:
\begin{equation}\label{eq:crowd_continues}
  N(0:T-1) = \sum_{t=1}^{t=T-1} N_{in}(t) + N(0).
\end{equation}
Considering that video frames are highly redundant, the inflow of most frames would be 0. Thus, we further simplify Eq. \ref{eq:crowd_continues} by inferring the inflow every $\uptau$ frames:
\begin{equation}\label{eq:crowd_discrete}
  N(0:T-1) \approx \sum_{k=1}^{k=T/\uptau} N_{in}^{\uptau}(k \times \uptau) + N(0),
\end{equation}
where $N_{in}^{\uptau}(t)$ is the inflow of frame $\mathbf{I}_{t}$ compared with $\mathbf{I}_{t-\uptau}$.

\begin{figure*}[thbp]
	\centering
	\includegraphics[width=0.98\textwidth]{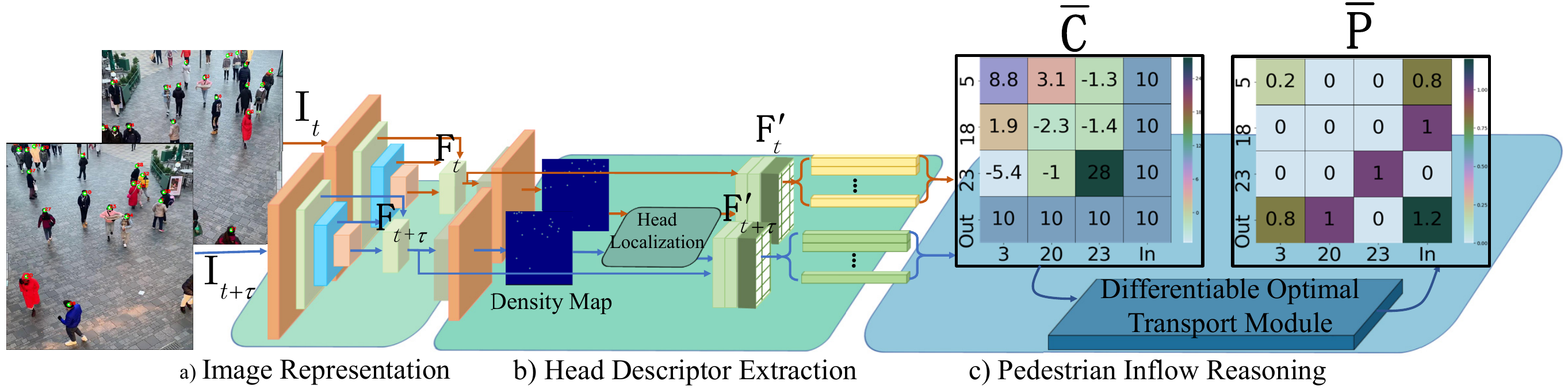}
	\vspace{-0.2cm}
	\caption{An overview of the DRNet architecture.
	DRNet is an end-to-end Video Individual Counting (VIC) framework, which takes the frames at $t$ and $t+\uptau$ as input and reasons the inflow at $t+\uptau$ compared with $t$. Based on the image representations obtained by a shared backbone network, two density maps can be obtained to guide the extraction of head descriptors.
	The pedestrian inflow in pair-wise images are given by reasoning their head descriptors to $\mathbf{\overline{P}}$.}
	\label{fig:frame}
	\vspace{-0.3cm}
\end{figure*}

\section{Method}

According to the formulation defined in Sec. \ref{Sec:Prolem}, our method should have the ability to count all pedestrians in the frame level and identify new pedestrians in a frame compared with its previous frame. We achieve this goal by designing an end-to-end network, called DRNet, to decompose video pedestrian counting as image-level pedestrian counting and cross-frame reasoning, which is composed of an image representation module, a head descriptor extraction module, and an inflow reasoning module.

\textbf{Image representation.} The image representation module maps two input images to feature maps in the high-level embedding space separately with a shared neural network.
As elucidated in Fig. \ref{fig:frame}, we sample a pair of images $\mathbf{I}_{t}$ and $\mathbf{I}_{t+\uptau}$ with a  time interval $\uptau$ from $\{\mathbf{I}_0, \mathbf{I}_1, \cdot\cdot\cdot, \mathbf{I}_{T-1}  \}$. The crowd images are transformed to corresponding multi-scale feature representations $\mathbf{F}_t$ and $\mathbf{F}_{t+\uptau} \in \mathbb{R}^{C\times H/4 \times W/4}$ with a backbone network, e.g., VGG-16 backbone \cite{simonyan2014very} and Feature Pyramid Network (FPN) \cite{lin2017feature} in this work.

\textbf{Head Descriptor Extraction.} Following existing density based crowd counting works~\cite{idrees2018composition, liu2019recurrent,gao2021domain,abousamra2020localization,wan2021generalized,song2021rethinking}, we use the feature maps from the image representation module to locate the head positions, which can be used to generate the descriptors (e.g., features) for each head center proposals. The details are elaborated in Sec. \ref{sec:head_descriptor}. The density map can also be used to generate the image level crowd count for the first frame directly in the testing phase.

\textbf{Pedestrian Inflow Reasoning.} Given the head descriptors of two frames from the head descriptor extraction module, the pedestrian inflow reasoning module targets on differentiation which subject is new in $\mathbf{I}_{t+\tau}$ compared with $\mathbf{I}_t$ by optimal transport. The details are elaborated in \cref{Sec:opt_solve}.

\subsection{Head Descriptor Extraction}\label{sec:head_descriptor}

As shown in the middle part of \cref{fig:frame}, the head description extraction module has two branches, one is head localization branch and the other is descriptors generation branch.
The localization branch packs several convolution and two deconvolution layers to map each image representation to a head density map, where the ground-truth head points are blurred with a Gaussian kernel $G(\sigma=4, \text{window size}=15)$. Thus, the coordinates of the local maximums in the density map are the head center proposals.

Denote the sets of head center proposals for frame $\mathbf{I}_{t}$ and $\mathbf{I}_{t+\uptau}$ by $\mathbf{p}_i^{t} := \{(h^{t}_{i},w^{t}_{i})\}_{i=1}^{M}$ and $\mathbf{p}_i^{t+\uptau} := \{(h^{t+\uptau}_{i},w^{t+\uptau}_{i})\}_{i=1}^{N}$  respectively. We first add some noises to augment these head proposals for improving the robustness,
 \begin{equation}
   \mathbf{p}_i^{t} \rightarrow \mathbf{p}_i^{t}+ \mathbf{z}_1, \quad \mathbf{p}_i^{t+\uptau}\rightarrow, \mathbf{p}_i^{t+\uptau}+\mathbf{z}_2
 \end{equation}
 where $\mathbf{z}_2$, $\mathbf{z}_2  \sim a \mathcal{N}(0,1)$. $a$ controls the noise level and is empirically set as 2. Then, the augmented proposals are expanded to $8\times8$ regions for pooling. Finally, two sets of head descriptors are extracted with the Precise RoI Pooling~\cite{jiang2018acquisition} on the final feature map $\mathbf{F}^{\prime}_{t}$ and $\mathbf{F}^{\prime}_{t+\uptau}$ (output by several convolution layers with $\mathbf{F}_{t}$ and $\mathbf{F}_{t+\uptau}$), denoted as $\mathcal{X}: = \{\mathbf{x}_1,... ,\mathbf{x}_{m} | \mathbf{x}_{i} \in \mathbb{R}^{D\times1}\}$  for $M$ subjects in $\mathbf{I}_t$ and $\mathcal{Y}: = \{\mathbf{y}_1,... ,\mathbf{y}_{n} | \mathbf{y}_{i} \in \mathbb{R}^{D\times1}\}$ for $N$ subjects in $\mathbf{I}_{t+\uptau}$. $D$ is the descriptor dimension and the default setting is 256.

\begin{figure}[t]
	\centering
	\includegraphics[width=0.47\textwidth]{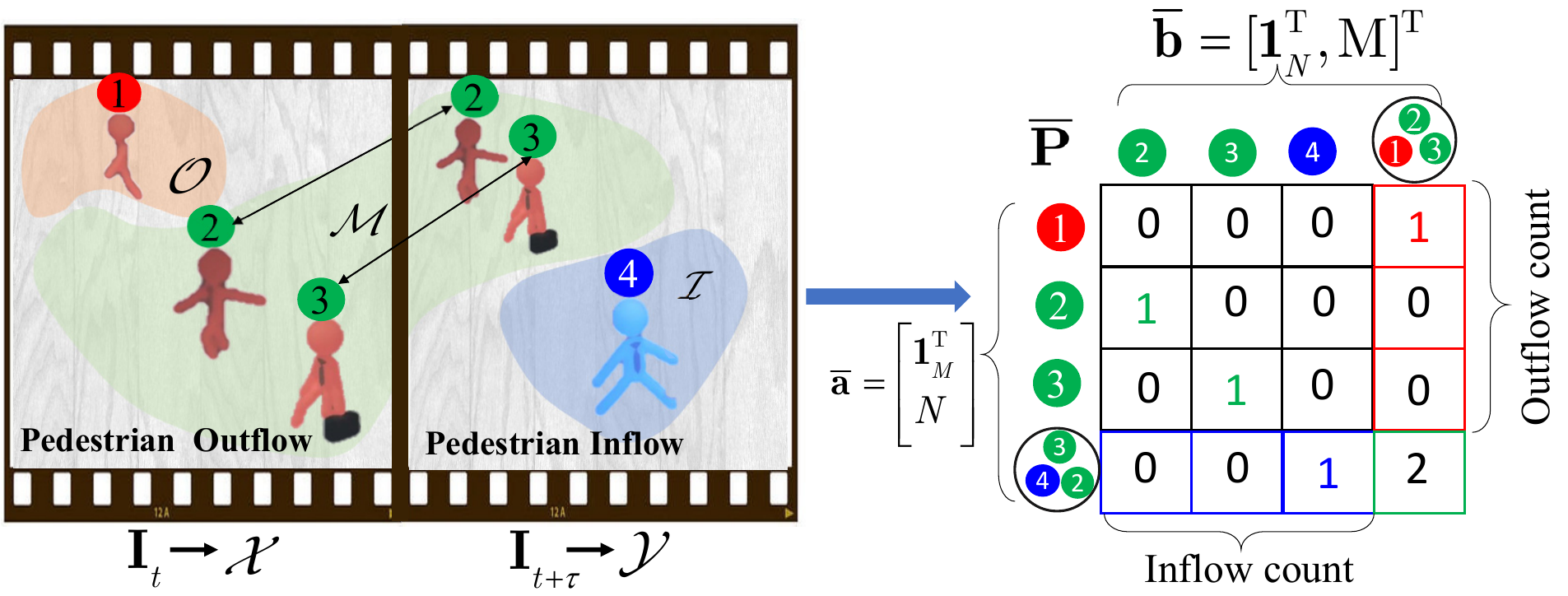}
	\vspace{-0.3cm}
	\caption{The sketch of the crowd transportation. With a proper solution of the transportation matrix $\mathbf{\overline{P}}$, the last row in  $\mathbf{\overline{P}}$ corresponds to the \textbf{inflow} for frame $t+\uptau$ that we want, \ie \ding{175}, the last column in  $\mathbf{\overline{P}}$ describes the \textbf{outflow} regarding frame $t$, \ie pedestrian \ding{172}, and matched pairs, \ie \ding{173}\ding{174}.}
	\label{fig:opt}
	\vspace{-0.3cm}
\end{figure}

\subsection{Pedestrian Inflow Reasoning}\label{Sec:opt_solve}
\noindent \textbf{Target}. Given the descriptors of the head center proposals obtained from the head descriptor extraction module (\cref{sec:head_descriptor}), the pedestrian inflow reasoning module is used for getting the inflow by associating the head center proposals.
As illustrated in Fig.~\ref{fig:opt}, given subjects in $\mathcal{X}$ and $\mathcal{Y}$, pedestrian inflow reasoning divides them into three categories,
 1) matched instance pairs $\mathcal{M}(\subseteq  \mathcal{X} \times \mathcal{Y})$ containing the people appearing in both $t$ and $t+\tau$ frame, 2) unmatched inflow instances $\mathcal{I}(\subseteq \mathcal{Y})$ containing the people only appearing in $t+\tau$ frame, and 3) unmatched outflow instances $\mathcal{O}(\subseteq  \mathcal{X})$ containing the people only appearing in $t$ frame.
 Thus, during the time duration $\uptau$, the size of set $\mathcal{I}$ is the number of inflow that we want to get, \ie the $N_{in}^{\uptau}(t+\uptau)$ in Eq. \ref{eq:crowd_discrete}. The outflow set $\mathcal{O}$ is a supernumerary output.

\noindent \textbf{Theoretic basis.} The problem above is a typical assignment problem, for which the Hungarian algorithm \cite{munkres1957algorithms} is a feasible solution. However, Hogarian algorithm requires a predefined distance threshold as classification basis, which would cost huge efforts to find the optimal threshold. More importantly, Hungarian algorithm is a non-differential process and only gives a hard matching (either zero or one). Hence it does not allow to optimize the image representation with the matching result. To avoid these issues, we chose the Optimal Transport (OT) theory \cite{peyre2019computational} to reason this assignment problem, which is widely used to plan the transportation between two target groups under some constrains. OT also provides a differentiable association process, which makes DRNet an end-to-end trainable network.

\noindent \textbf{The reasoning for inflow.}
Inspired by the solutions used in the graph matching and the key-point matching tasks \cite{detone2018superpoint,sarlin2020superglue}, the pedestrian inflow in our paper can be obtained by solving a augmented Kantorovich's OT problem,

\begin{equation}\label{Eq:OP_define}
\mathrm{L}_{\mathbf{\overline{C}}}(\mathbf{\overline{a}}, \mathbf{\overline{b}})=\min _{\mathbf{\overline{P}} \in \mathbf{U}(\mathbf{\overline{a}}, \mathbf{\overline{b}})} \sum_{i \in \llbracket M+1 \rrbracket, j \in \llbracket N+1 \rrbracket} \mathbf{\overline{C}}_{i, j} \mathbf{\overline{P}}_{i, j},
\end{equation}
 where $\mathbf{\overline{P}}$ is a transportation matrix. As depicted in \cref{fig:opt}, its element $\mathbf{\overline{P}}_{i,j} (0 \leq i \textless M, 0 \leq j \textless N)$ represents the probability of the $i$th pedestrian in preceding frame is associated with the $j$th people in the current frame. Note that the augmented row $\mathbf{\overline{P}}_{(M+1,j)} (0 \leq j \textless N)$ is defined to collect the pedestrian inflow  (\ie inflow container),
\begin{equation} \label{Eq:inflow_solution}
N_{in}^{\tau}(t+\tau) = \sum_{j=1}^{N} \mathbf{\overline{\mathbf{P}}}_{(M+1,j)},
\end{equation}
Eq. \ref{Eq:inflow_solution} shows pedestrian inflow reasoning is a transportation reasoning problem, predicting the probabilities of staying in the scene and matching to the inflow/outflow. Similarly, the supernumerary outflow can be collected by the augmented column $\mathbf{\overline{P}}_{(i,N+1)} (0 \leq i \textless M)$ (\ie outflow container).
$\mathbf{\overline{C}}$ is the cost matrix and actually is a similarity matrix calculated with descriptor sets $\mathcal{X}$, $\mathcal{Y}$ and two augmented bins in this paper,
\begin{equation}\label{Eq:OT_new_cost}
\mathbf{\overline{C}}=\left[\begin{array}{l}
\mathbf{C}_{M,N} \quad c_{M,1} \\
c_{1,N} \quad \quad c_{1,1}
\end{array}\right] \in \mathbb{R}^{(M+1) \times (N+1)},
\end{equation}
where $c_{M,1}, c_{1,N}$ and $c_{1,1}$ are expanded by a learnable parameter $c$, and act as thresholds to discriminate whether or not a person existed in both two frames. $c_{M,1}$ and $c_{1,N}$ represent the possibilities for the person matched to the outflow container and inflow container, respectively.

The $(i, j)$-th element $\mathbf{C}_{i,j}$ in matrix $\mathbf{C_{M,N}}$ uses the features of the $i$-th person at frame $t$ and the $j$-th person at frame $t+\uptau$ to measure their similarity as follows:
\begin{equation}
\mathbf{C}_{i,j}=\mathbf{x}_{i}^{\top}{\mathbf{y}}_{j}, \forall(i,j) \in \llbracket M \rrbracket \times \llbracket N \rrbracket.
\end{equation}

The $\mathbf{\overline{U}}(\mathbf{\overline{a}}, \mathbf{\overline{b}})$ in Eq. \ref{Eq:OP_define} is a discrete measure with respect to probability vectors $\mathbf{\overline{a}}$ and $\mathbf{\overline{b}}$,
\begin{equation}\label{Eq:OT_new_constrains}
\mathbf{\overline{U}}(\mathbf{\overline{a}}, \mathbf{\overline{b}}) \stackrel{\text { def. }}=\left\{\mathbf{\overline{P}}: \mathbf{\overline{P}} \mathbb{1}_{N}=\mathbf{\overline{a}} \text { and } \mathbf{\overline{P}}^{\mathrm{T}} \mathbb{1}_{M}=\mathbf{\overline{b}}\right\}.
\end{equation}
where $\mathbb{1}$ is the
column vector of ones. To solve Eq. \ref{Eq:OP_define}, we need to give a reasonable prior probability vector $\mathbf{\overline{a}}$ (for $\mathcal{X}$) and $\mathbf{\overline{b}}$ (for $\mathcal{Y}$) in advance. Actually, instances in $\mathcal{X}$ or $\mathcal{Y}$ can be regarded with the same probability to be matched as they are of equal importance in pedestrian counting. Hence their mass are all set as 1. As for inflow row and outflow column, their masses are respectively defined as $N$ and $M$ so as to create equal constrains. Finally, two histogram vectors $\mathbf{\overline{a}} = [\mathbb{1}^{\top}_{M} \quad N]^{\top}$ and $\mathbf{\overline{b}} = [\mathbb{1}^{\top}_{N} \quad M]^{\top}$ are used to solve $\mathbf{\overline{P}}$.

The overall objective of OT problem is to find matrix $\mathbf{\overline{P}}$ reasoning the pedestrians in $\mathcal{X}$ towards $\mathcal{Y}$ so that $\sum \overline{\mathbf{P}} \times \overline{\mathbf{C}}$ is the maximum. In fact, this is a linear programming problem (Eq. \ref{Eq:OP_define}) with $N+M+2$ constraints (Eq. \ref{Eq:OT_new_constrains}).

\noindent \textbf{Differentiable and approximate solution of OT (DOT).}
The final step of DRNet is to use a differentiable algorithm to solve the assignment weight matrix $\mathbf{\overline{P}}$, so that we can optimize the head descriptors with the assignment results. The standard solution of the original Kantorovich's OT problem has high time complexity and it is hard to solve. An approximated solution of the regularized Kantorovich's OT problem in  \cite{peyre2019computational} is given as,
 \begin{equation} \label{Eq:14}
\mathbf{\overline{P}}_{i, j}=\mathbf{u}_{i} \mathbf{K}_{i, j} \mathbf{v}_{j},
\end{equation}
where $\mathbf{K}_{i, j}=e^{-\mathbf{\overline{C}_{i,j}}/\sigma}$, $\sigma$ is the regularization coefficient and we set it as 1, and the vectors $\mathbf{u}$ and $\mathbf{v}$ are variables, which are solved by the Sinkhorn algorithm \cite{sinkhorn1967concerning,cuturi2013sinkhorn}. According to the Eq. \ref{Eq:OT_new_constrains} and \ref{Eq:14}, we can update $\mathbf{u}$ and $\mathbf{v}$ by alternately iterating the following two equations,
\begin{equation}\label{Eq:15}
\mathbf{u}^{(\ell+1)} \stackrel{\text { def. }}{=} \frac{\mathbf{a}}{\mathbf{K v}^{(\ell)}} \quad \text { and } \quad \mathbf{v}^{(\ell+1)} \stackrel{\text { def. }}{=} \frac{\mathbf{b}}{\mathbf{K}^{\mathrm{T}} \mathbf{u}^{(\ell+1)}},
\end{equation}
where $\mathbf{a} = \frac{\mathbf{\overline{a}}}{MN}$ and $\mathbf{b} = \frac{\mathbf{\overline{b}}}{MN}$ are, respectively, the normalized versions $\mathbf{\overline{a}}$ and $\mathbf{\overline{b}}$. $\mathbf{v}^{(0)}$ is initialized by $\mathbb{1}_N$. $l$ is the iterations and the default setting is 100. Sarlin \etal~\cite{sarlin2020superglue} provide a fast-speed computation of the Sinkhorn algorithm and the time consumption is extremely low with 100 iteration, which accounts for approximately 3\% training time in DRNet. Eq. \ref{Eq:15} reveals that inferring $\mathbf{\overline{P}}$ is a completely differential process.

\subsection{Loss Functions}
Two loss functions are used in this framework: 1) A standard MSE loss supervises the density prediction task as widely used in the image-level crowd counting \cite{zhang2016single,sam2017switching,li2018csrnet}.
 2) a matching loss $\mathcal{L}_{match} = \mathcal{L}_{p} + \mathcal{L}_{h}$ can maximize the likelihood probability for positive samples and minimize the likelihood probability for hard negative samples,

\begin{gather}
 \mathcal{L}_{p}  = -\sum_{(i,j) \in \mathop{\arg}\limits_{i,j} (\mathbf{\overline{P}}_g==1)} \log\mathbf{\overline{P}}_{i,j}, \label{Eq:matching loss1}\\
 \mathcal{L}_{h} = -\sum_{(i,j) \in \mathop{\arg}\limits_{i,j} (\mathbf{\overline{P}}_h==1)} \log(1-\mathbf{\overline{P}}_{i,j}), \label{Eq:matching loss2}
\end{gather}
where the $\mathbf{\overline{P}}_g \in \{0,1\}^{(m+1)\times (n+1)}$ is ground truth assignment matrix, which is generated by the the annotated association labels. $\mathop{\arg}\limits_{i,j}(\mathbf{\overline{P}}_g==1)$ returns the indexes of the elements in $\mathbf{\overline{P}}_{g}$ with the value as 1. Eq. \ref{Eq:matching loss1} enforces the feature presentations of the same person to be similar in the different frames. Eq. \ref{Eq:matching loss2} is designed to enlarge the distance between a person and his/her hard samples. Its ground truth matrix $\mathbf{\overline{P}}_{h}$ is adaptively generated by finding the hardest sample for each instance according to prediction $\mathbf{\overline{P}}$.

\section{Experiments}
\subsection{Datasets}
We find two benchmark datasets, i.e., CroHD \cite{sundararaman2021tracking} and SenseCrowd \cite{li2021video}, are suitable for VIC. Both of them have annotations for head locations and associations of pedestrians. \cref{Table:datasets} describes the detailed statistics of them. These two datasets contain diverse spots, especially congested spots.
For CroHD, we use four videos for training and validation, and five videos for testing as the official splitting.
For SenseCrowd, all video clips are randomly divided into training ($50\%$), validation ($10\%$), and testing ($40\%$).
Besides, since SenseCrowd is large-scale and contains diverse scenes, we further manually label all videos with different scene labels for a more comprehensive analysis.

\begin{table}[htbp]	
\centering
\small
	\setlength{\tabcolsep}{0.7mm}{\begin{tabular}{cIc|c|c|c|c}
		\whline
		Dataset   & Videos & Frames &Head labels &Pedestrians&Time (s)   \\
		\hline
		CroHD     & 9     & 11,463 &2,276,838 &5,230  &498        \\
        \hline
        SenseCrowd& 634    & 62,938 &2,344,276 &43,178  &12,588      \\
		\whline		
	\end{tabular}}
	\vspace{-0.2cm}
\caption{Summary of the video datasets for pedestrian counting.}
\label{Table:datasets}
\end{table}

\subsection{Evaluation Metrics}
Follow existing counting tasks (e.g. crowd counting \cite{zhang2016single}, vehicle counting \cite{8648370}), we use Mean Absolute Error (MAE) and Mean Square Error (MSE) for evaluation. Different from image-level crowd counting, we calculate them based on the whole video pedestrian count with the same person only counted once. Besides, we also use a Weighted Relative Absolute Errors (WRAE) to balance the performance on videos with different lengths and pedestrian numbers,
\begin{equation}\label{eq:wrae}
\small
  WRAE = \sum_{i=1}^{K} \frac{T_{i}}{\sum_{j=1}^{K} T_{j}}  \frac{|N_i-\hat{N}_i|}{N_i} \times 100\%,
\end{equation}
where $N_i$ and $\hat{N}_i$ respectively represent the annotated and estimated pedestrian number for the $i$-th test video. $K$ is the total number of videos. $T_{i}$ is the total number of frames for the $i$-th video.  
Since our method involves association within two frames, we further use the Mean Inflow/Outflow Absolute Error (MIAE/MOAE) to reflect the association quality. (More descriptions are given in the Supplementary.)

\begin{table*}[htbp]	
  \small
  \centering
  \setlength{\tabcolsep}{0.75mm}{\begin{tabular}{cIcIccIcccccIccc}
    \whline
    \multirow{3}{*}{Methods}& \multicolumn{3}{cI}{\emph{val} set} & \multicolumn{5}{cI}{Counting results in five testing scenes} &\multicolumn{3}{c}{Metrics on test set}\\
    \cline{2-12}
    & CroHD01&\multirow{2}{*}{MIAE$\downarrow$}&\multirow{2}{*}{MOAE$\downarrow$}& CroHD11 &CroHD12 &CroHD13&CroHD14&CroHD15 &\multirow{2}{*}{MAE$\downarrow$} &\multirow{2}{*}{MSE$\downarrow$}&\multirow{2}{*}{MRAE(\%)$\downarrow$}\\
    & \underline{85}&   &  &   \underline{133} &\underline{737} &\underline{734}&\underline{1040}&\underline{321} & &&\\
    \whline

    PHDTT \cite{phdtt}*   &183&9.1&18.1 & 380 &4530 &5528 &1531&1648   &2130.4&2808.3&401.6\\
    FairMOT \cite{zhang2021fairmot} & 214 &6.0 &6.7  &\textbf{144} &1164 &1018 &632&472  &256.2&300.8&44.1 \\
    HeadHunter-T \cite{sundararaman2021tracking} &145 &\textbf{5.1}& 6.2  &198 &636 &219 &458 &324 &253.2&351.7&32.7\\
    \whline
    LOI  \cite{zhao2016crossing}     &\textbf{65.5} &-&-  & 72.4 &\textbf{493.1} &275.3 &409.2&189.8   &305.0&371.1&46.0 \\
    \whline
     DRNet& 113.0 &6.1&\textbf{4.4}   &164.6&1075.5&\textbf{752.8} &\textbf{784.5} &\textbf{382.3}       &\textbf{141.1} &\textbf{192.3} &\textbf{27.4}\\
    \whline
  \end{tabular}}
  \vspace{-0.3cm}
  \caption{Video individual counting performance on CroHD dataset. *: the code is not available and results come from the submission of MOTChallenge \cite{MOTchallenge}. The \underline{underline} fonts represent ground truth. `-' means the metrics cannot be calculated with the corresponding algorithms. Overall, DRNet obtains more accurate counts than the tracking methods \cite{zhang2021fairmot,phdtt,sundararaman2021tracking} and cross-line counting method \cite{zhao2016crossing}. }
  \label{tab:CroHead}
  \vspace{-0.4cm}
\end{table*}

\begin{table*}[htbp]	
  \small
  \centering
  \setlength{\tabcolsep}{2.75mm}{\begin{tabular}{cIcccccIccccc}
    \whline

    \multirow{2}{*}{Methods}& \multicolumn{5}{cI}{Overall} & \multicolumn{5}{c}{Density (for MAE)} \\
    \cline{2-11}
    & MAE$\downarrow$& MSE$\downarrow$&MRAE(\%)$\downarrow$&MIAE$\downarrow$& MOAE$\downarrow$&D0&D1&D2&D3&D4\\
    \whline
    FairMOT \cite{zhang2021fairmot}  &35.4 &62.3&48.9&4.9&4.4  &13.5&22.4&67.9&84.4&145.8 \\
    HeadHunter-T \cite{sundararaman2021tracking} &30.0 &50.6&38.6&4.0&4.1  &11.8&25.7&56.0&92.6&131.4 \\

    \whline
    LOI  \cite{zhao2016crossing}  &24.7 &33.1&37.4&-&-  &12.5&25.4&39.3&\textbf{39.6}&86.7 \\
    \whline
     DRNet &\textbf{12.3}&\textbf{24.7}&\textbf{12.7} & \textbf{1.98} & \textbf{2.01}  &\textbf{4.1}&\textbf{8.0}&\textbf{23.3}&50.0&\textbf{77.0}\\
     \whline
  \end{tabular}}
  \vspace{-0.2cm}
  \caption{Video individual counting performance on SenseCrowd dataset. $D0\sim D4$ respectively indicate five pedestrian density ranges: $[0,50),[50,100),[100,150),[150,200), \geq 200$. More results about the performance with different locations, day\&night, indoor\&outdoor are reported in Appendix.}
  \vspace{-0.2cm}
  \label{tab:SenseCrowd}
\end{table*}

\subsection{Implementation Details}
\textbf{Training details:} For efficient training, the time interval for each image pair is randomly sampled from range $2s\sim8s$ to guarantee the pair contain both matched and unmatched samples. For data augmentation, we use the random horizontally flipping, scaling ($0.8\times\sim1.25\times$), and cropping (768$\times$1024) strategies. The learning rate is initialized to $5e-5$ except the $1e-2$ for $c$, and they are updated by a step decay strategy with 0.95 rate at each epoch. Adam \cite{kingma2014adam} algorithm is adopted to optimize the framework. The VGG-16 backbone is initialized with the pre-trained weights on ImageNet \cite{krizhevsky2012imagenet}. The model is built upon the Pytorch framework \cite{paszke2019pytorch} and implemented on an TITAN RTX GPU (24G memory) with batchsize 4.

\textbf{Testing details:} In the testing phase, the time interval $\uptau$ is fixed as 3s except for the time interval analysis presented in \cref{sec:time}.  

\subsection{Overall Comparison}
 \textbf{Comparison methods:}
To evaluate the effectiveness of our method, we adapt some relevant works to the individual counting task for comparison. These works are classified into two categories. 1) Tracking-based: the tracking results of three advanced MOT methods, i.e., HeadHunter-T \cite{sundararaman2021tracking}, FairMOT \cite{zhang2021fairmot}, and PHDTT \cite{phdtt} 
are employed to estimate the pedestrian flow by counting their tracks. For the CroHD dataset, we first try to directly use the tracking results provided in MOTChallenge \cite{MOTchallenge}, but the errors (MAE/MSE) is very large. By lowering the frame rate to 0.33 FPS for FairMOT and 1 FPS for HeadHunter-T, trackingh-based methods produce their best results. For the SenseCrowd dataset, we use the official public code to get the tracking results (PHDTT is omitted for SenseCrowd as the code is not available). 2) Density-based: the representative cross-line crowd counting method LOI \cite{zhao2016crossing} is re-implemented \footnote{The official source code is not available.} and assessed with our evaluation system.

\textbf{Results on CroHD:} To the best of our knowledge, we are the first to conduct video pedestrians counting on such a congested dataset. \cref{tab:CroHead} outlines the pedestrian numbers in each scene of the testing set as well as three metrics on all videos. DRNet outperforms all tracking-based methods and cross-line method with an obvious improvement. The overall MAE and MSE are lowered to 141.1 and 192.3, respectively. The errors of some tracking methods are more than tenfold those of our MAE and MSE if we don't change  its FPS in testing. 
The reason is that wrong associations are normal in the extremely congested scenes, which would accumulate to following associations and make tracks lose targets. In the following frames, new tracks would be added for existing pedestrians frequently. However, the reasoning error will not influence the following matching in DRNet own to the decomposition. Besides, DRNet also requires less association steps, which are only conducted in sampled frames.
This also can be used to explain why DRNet surpasses other methods substantially despite the MIAE and MOAE being only slightly better than other methods.
Notably, the numbers estimated by LOI~\cite{zhao2016crossing} are all fewer than the GTs, which meets the expectation and verifies it is not a stable method to count all pedestrians in complex scenarios.

\textbf{Results on SenseCrowd:} \cref{tab:SenseCrowd} shows the results on SenseCrowd. DRNet makes the best predictions on the overall dataset as well as the different density subsets (D0$\sim$D4), demonstrating its effectiveness. Especially, the errors are much smaller than those in CroHD since SenseCrowd is sparser. Our MRAE ($12.7\%$) is relatively low, making it possible to be deployed in the future. The overall counting performance would be better if the assignment and head localization accuracies are further improved.
\begin{table*}[htbp]	
  \centering
  \small
  \setlength{\tabcolsep}{0.8mm}{\begin{tabular}{cIcIcccccIccc}
    \whline
    \multirow{3}{*}{Investigated}&\multirow{3}{*}{Settings} &\multicolumn{5}{cI}{Counting results in five testing scenes} &\multicolumn{3}{c}{Metrics on test set} \\

     \cline{3-10}
     && CroHD11 &CroHD12 &CroHD13&CroHD14&CroHD15 &\multirow{2}{*}{MAE$\downarrow$} &\multirow{2}{*}{MSE$\downarrow$} &\multirow{2}{*}{MRAE(\%)$\downarrow$}\\
    && \underline{133} &\underline{737} &\underline{734}&\underline{1040}&\underline{321} \\
    \cline{3-9}

    \hline
    \multirow{2}{*}{Association methods}
    & Tracking \cite{sundararaman2021tracking}&284&  1364& 1435& 1917& 539 &526.4&604.8&87.7\\

    & Hungarian\cite{munkres1957algorithms}
    & 129&421&332&331&185&       313.4&395.6&45.4\\
    \hline
    \multirow{1}{*}{Head Proposals} &GT           &176.9&1357.0&1118.0&1029.6&518.6  &251.2&338.5&54.1   \\
    \hline
    \multirow{1}{*}{Hard Negative Pair: $\mathcal{L}_h$} &W/O       & 151.7&1213.3&779.0&768.9&456.8   &189.4& 253.4 &38.2\\
    \whline
     DRNet  &DOT+GT+Pred+$\mathcal{L}_{match}$ &164.6&1075.5&752.8 &784.5 &382.3     &141.1 &192.3   &27.4\\
    \whline
  \end{tabular}}
  \vspace{-0.2cm}
  \caption{Ablation study for DRNet. ``Tracking~\cite{sundararaman2021tracking}'' denotes the data association method of \cite{sundararaman2021tracking}. ``GT'' means only using ground truth as head proposals during training. The \underline{underline} fonts represent ground truth.
All methods are with the same time interval in matching.}
  \label{tab:ablation}
  \vspace{-0.2cm}
\end{table*}

\subsection{Ablation Study}
\textbf{Assignment Methods:} Besides the Differential Optimal Transport (DOT) in this paper, we also consider two heuristic matching methods to achieve pedestrians association from a pair of frames: Data association in MOT \cite{sundararaman2021tracking,zhang2021fairmot} and the Hungarian algorithm \cite{munkres1957algorithms}. In the ablation study, we first train the network with the full DRNet and then replace the DOT module with other association methods for testing. \cref{tab:ablation} shows that the association method in MOT has the largest error, whereas the Hungarian algorithm \cite{munkres1957algorithms} can make a better matching performance with an appropriate threshold. In fact, DOT is a differential version of the Hungarian algorithm, it makes the best counting results with end-to-end learning.

\textbf{Head Proposals:} During training, we can use either the combination of predicted head proposals and GT points or only the GT points. Here, we analyse the contribution of the predicted head proposals. The results in the last row and the third last row in \cref{tab:ablation} show that the predicted head centers at training stage substantially improves the counting performance, decreasing MAE and MRAE by $78.0\%$ and  $26.7\%$, respectively.

\textbf{Hard Negative Pair Loss $\mathcal{L}_{h}$:}
Since we design $\mathcal{L}_{h}$ in Eq. \ref{Eq:matching loss2} to further enlarge the feature distance of different people, thus we conduct an experiment to discuss how much gain this design brings in. Comparison between the last two rows of \cref{tab:ablation} shows that $\mathcal{L}_{h}$ makes a significant promotion. Take the MRAE for an example, it further drops to $27.4\%$ by including hard negative pairs loss.

\begin{figure}[htbp]
\centering
\begin{minipage}[t]{0.23\textwidth}
\centering
\includegraphics[width=0.98\textwidth]{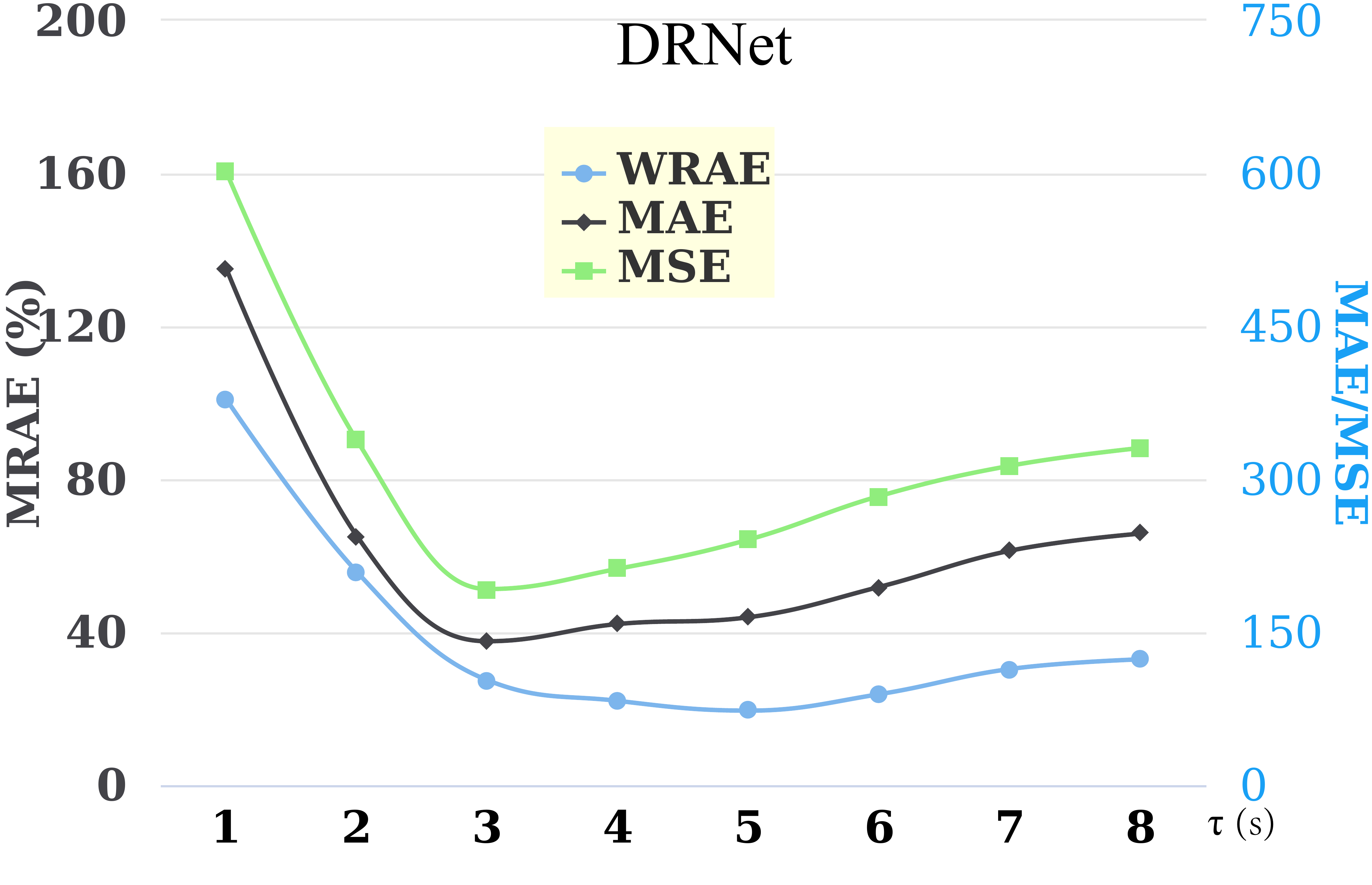}
\end{minipage}
\begin{minipage}[t]{0.23\textwidth}
\centering
\includegraphics[width=0.98\textwidth]{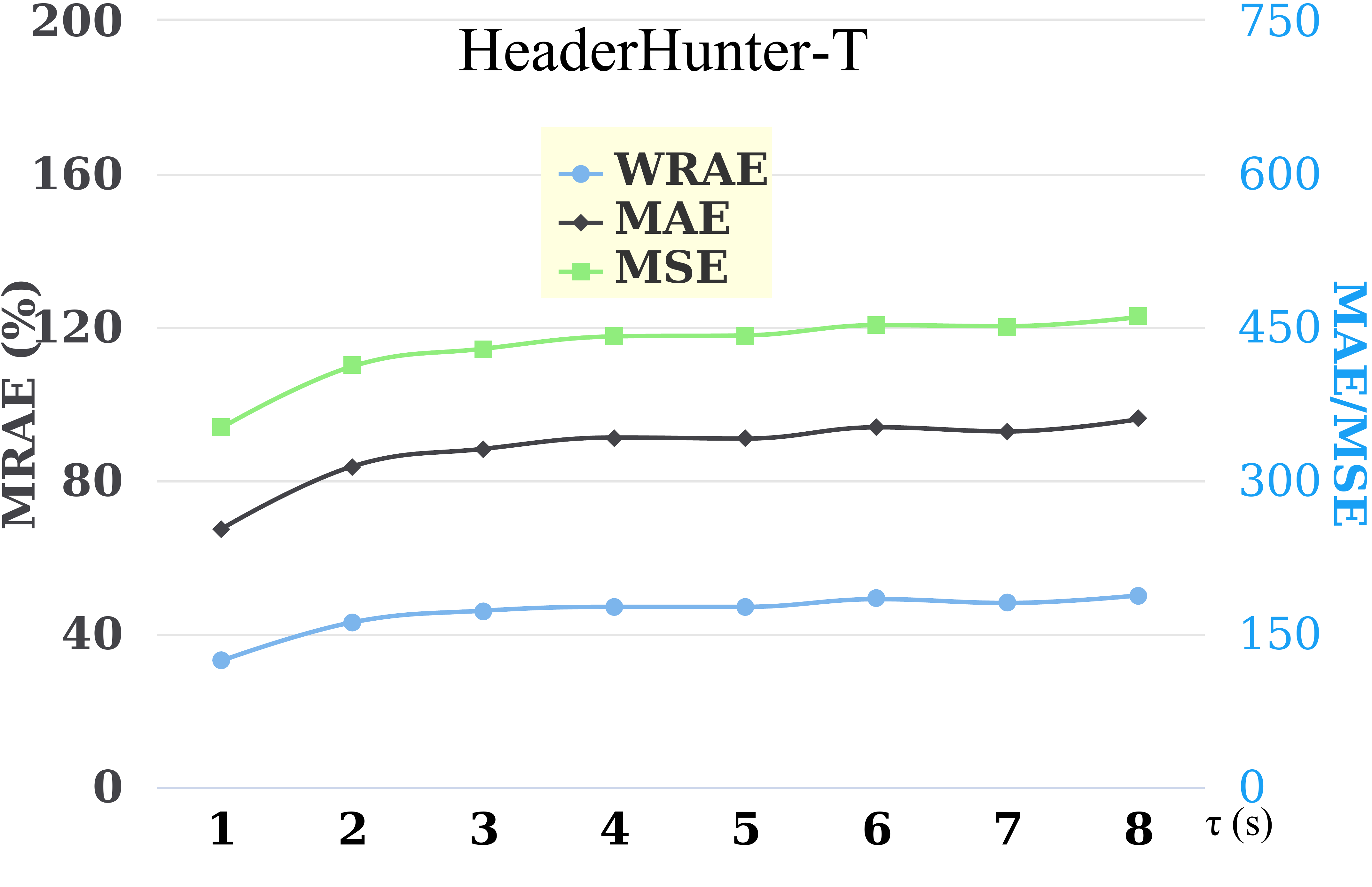}
\end{minipage}
\vspace{-0.2cm}
\caption{Errors MRAE, MAE, and MSE (Y-axis) on CroHD for different counting intervals (X-axis). DRNet achieves promising performance with a relatively large time slot (e.g., 3-4 seconds), while HeadHunter-T tends to rely on successive frames.}
\label{fig:time_slot}
\vspace{-0.3cm}
\end{figure}

\textbf{Influence of time interval:} \label{sec:time}
The above results of DRNet  are tested with a fixed time slot $\uptau$. Here, we investigate the influence of $\uptau$ to our counting performance on the CroHD dataset. Besides, we also conduct experiments on HeadHunter-T \cite{sundararaman2021tracking} at the same time intervals for comparison. As shown in \cref{fig:time_slot}, DRNet can achieve excellent individual counting performance with a suitable time interval (e.g., 3$\sim$4s), which also significantly decreases computation cost since less reasoning is required. However, the error rates for tracking-based HeadHunter-T steadily increase with the increase of time interval. Combined with the comparison in Tab. \ref{tab:CroHead}, we can conclude that DRNet can achieve much better performance when compared with the tracking-based methods in terms of both accuracy and efficiency.

\subsection{Qualitative Results}
\begin{figure}
	\centering
	\includegraphics[width=0.47\textwidth]{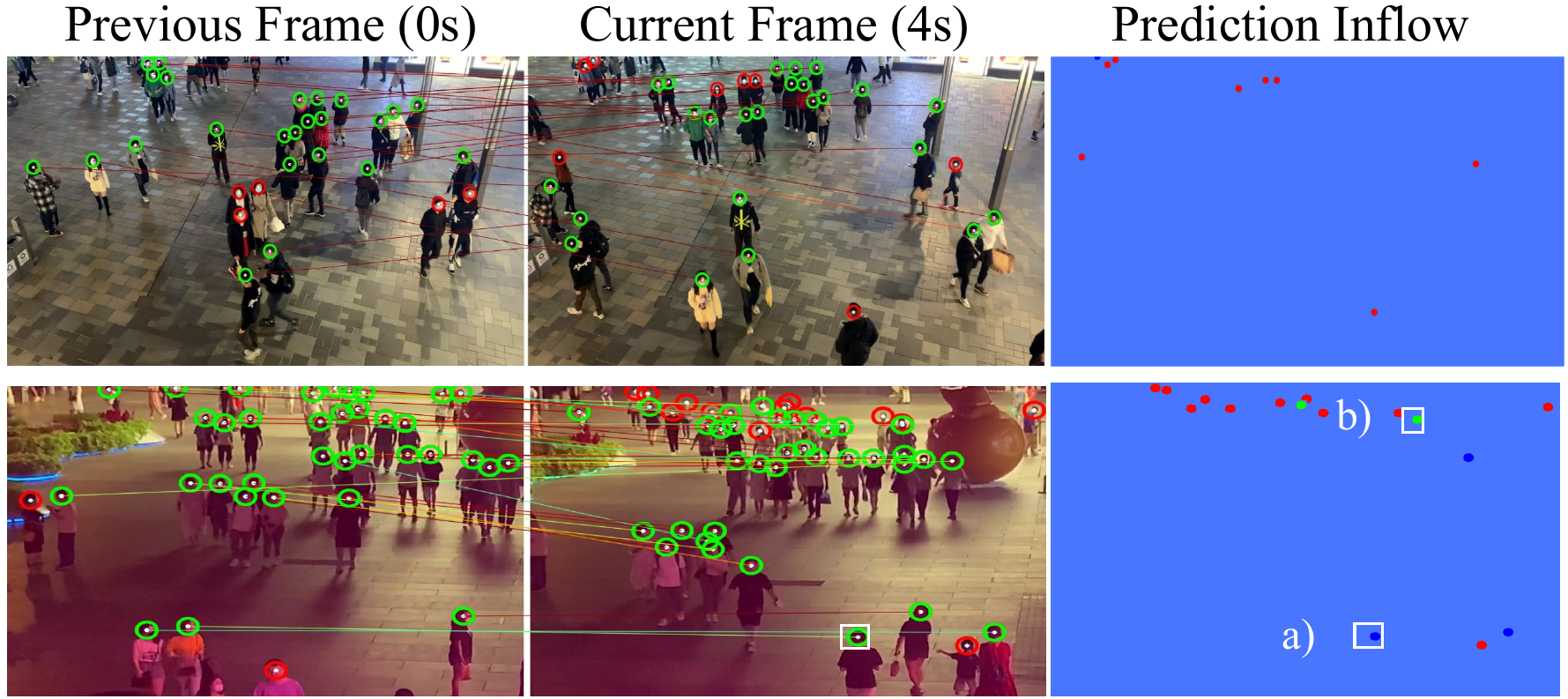}
	\vspace{-0.3cm}
	\caption{ Visualization samples in night scenes. The green and red circles in the 1st and 2nd columns denote matched and unmatched pedestrians, respectively. 
	The {\color{red}{red}}, {\color{blue}{blue}}, and {\color{green}{green}} points in 3rd column respectively denote correctly identified inflow, missed inflow, and over-counted new pedestrians, respectively.} 
	\label{fig:result}
	\vspace{-0.5cm}
\end{figure}

 \cref{fig:result} visualizes the head proposals of the reasoned inflow outputs from DRNet on two night scenes, which is challenging for counting.
 Overall, DRNet makes a precise reasoning in moderately dark scenario as shown in the first row. 
 However, there are also some failed cases in the more complicated scene as shown in Column 3 of the second row. For instance, a) the wrong assignment would decrease the pedestrian inflow number (the blue point with white box in last column), or b) over-claim an existing pedestrian as inflow (the green point with white box in last column).

\noindent \emph{Limitations and potential negative societal impact are disscussed in the supplementary.}

\section{Conclusion}
We study the video individual counting task by decomposing all individuals in the video to the initial individuals at the first frame and a set of new individuals at each following frame, which is a new direction for video level crowd counting. An end-to-end learnable network named DRNet is proposed to achieve this idea by estimating the pedestrians density map and reasoning the inflows of frame-pairs with the differential optimal transport. 
Experiments on two datasets with congested and diverse scenes demonstrate the effectiveness of DRNet over competitive baselines. Since DRNet only reasons the association on sampled frame pairs with a large interval, the computational efficiency is also attractive. We believe this direction will make a significant promotion for crowd analysis and attract more research's interests in video individual counting and crowd analysis.

\textbf{Acknowledgment} This work was supported by the National Natural Science Foundation of China under Grant U21B2041 and U1864204. Wanli Ouyang was supported by the Australian Research Council Grant DP200103223, Australian Medical Research Future Fund MRFAI000085, and CRC-P Smart Material Recovery Facility (SMRF) – Curby Soft Plastics.


{\small
\bibliographystyle{ieee_fullname}
\bibliography{egbib}

\begin{thebibliography}{10}\itemsep=-1pt

\bibitem{MOTchallenge}
Motchallenge.
\newblock [Online].
\newblock \url{https://motchallenge.net}.

\bibitem{abousamra2020localization}
Shahira Abousamra, Minh Hoai, Dimitris Samaras, and Chao Chen.
\newblock Localization in the crowd with topological constraints.
\newblock {\em arXiv preprint arXiv:2012.12482}, 2020.

\bibitem{albiol2001real}
Antonio Albiol, Inmaculada Mora, and Valery Naranjo.
\newblock Real-time high density people counter using morphological tools.
\newblock {\em IEEE TITS}, 2(4):204--218, 2001.

\bibitem{sam2020locate}
Deepak Babu~Sam, Skand~Vishwanath Peri, Mukuntha Narayanan~Sundararaman, Amogh
  Kamath, and Venkatesh~Babu Radhakrishnan.
\newblock Locate, size and count: Accurately resolving people in dense crowds
  via detection.
\newblock {\em IEEE TPAMI}, 2020.

\bibitem{bai2020adaptive}
Shuai Bai, Zhiqun He, Yu Qiao, Hanzhe Hu, Wei Wu, and Junjie Yan.
\newblock Adaptive dilated network with self-correction supervision for
  counting.
\newblock In {\em CVPR}, pages 4594--4603, 2020.

\bibitem{barandiaran2008real}
Javier Barandiaran, Berta Murguia, and Fernando Boto.
\newblock Real-time people counting using multiple lines.
\newblock In {\em 2008 Ninth International Workshop on Image Analysis for
  Multimedia Interactive Services}, pages 159--162. IEEE, 2008.

\bibitem{bera2014adapt}
Aniket Bera, Nico Galoppo, Dillon Sharlet, Adam Lake, and Dinesh Manocha.
\newblock Adapt: real-time adaptive pedestrian tracking for crowded scenes.
\newblock In {\em ICRA}, pages 1801--1808. IEEE, 2014.

\bibitem{bescos2003dct}
Jes{\'u}s Besc{\'o}s, Jos{\'e}~M Men{\'e}ndez, and Narciso Garc{\'\i}a.
\newblock Dct based segmentation applied to a scalable zenithal people counter.
\newblock In {\em ICIP}, volume~3, pages 14--17. IEEE, 2003.

\bibitem{chan2008privacy}
Antoni~B. Chan, Zhang-Sheng~John Liang, and Nuno Vasconcelos.
\newblock Privacy preserving crowd monitoring: Counting people without people
  models or tracking.
\newblock In {\em CVPR}, 2008.

\bibitem{chaudhary2021video}
Deevesh Chaudhary, Sunil Kumar, and Vijaypal~Singh Dhaka.
\newblock Video based human crowd analysis using machine learning: a survey.
\newblock {\em Computer Methods in Biomechanics and Biomedical Engineering:
  Imaging \& Visualization}, pages 1--19, 2021.

\bibitem{cong2009flow}
Yang Cong, Haifeng Gong, Song-Chun Zhu, and Yandong Tang.
\newblock Flow mosaicking: Real-time pedestrian counting without scene-specific
  learning.
\newblock In {\em CVPR}, pages 1093--1100. IEEE, 2009.

\bibitem{cuturi2013sinkhorn}
Marco Cuturi.
\newblock Sinkhorn distances: Lightspeed computation of optimal transport.
\newblock {\em NeurIPS}, 26:2292--2300, 2013.

\bibitem{detone2018superpoint}
Daniel DeTone, Tomasz Malisiewicz, and Andrew Rabinovich.
\newblock Superpoint: Self-supervised interest point detection and description.
\newblock In {\em CVPRW}, pages 224--236, 2018.

\bibitem{dong2020scale}
Zihao Dong, Ruixun Zhang, Xiuli Shao, and Yumeng Li.
\newblock Scale-recursive network with point supervision for crowd scene
  analysis.
\newblock {\em Neurocomputing}, 384:314--324, 2020.

\bibitem{fang2019locality}
Yanyan Fang, Biyun Zhan, Wandi Cai, Shenghua Gao, and Bo Hu.
\newblock Locality-constrained spatial transformer network for video crowd
  counting.
\newblock In {\em ICME}, pages 814--819. IEEE, 2019.

\bibitem{gao2020learning}
Junyu Gao, Tao Han, Yuan Yuan, and Qi Wang.
\newblock Learning independent instance maps for crowd localization.
\newblock {\em arXiv preprint arXiv:2012.04164}, 2020.

\bibitem{gao2021domain}
Junyu Gao, Tao Han, Yuan Yuan, and Qi Wang.
\newblock Domain-adaptive crowd counting via high-quality image translation and
  density reconstruction.
\newblock {\em IEEE Transactions on Neural Networks and Learning Systems},
  2021.

\bibitem{gao2020feature}
Junyu Gao, Yuan Yuan, and Qi Wang.
\newblock Feature-aware adaptation and density alignment for crowd counting in
  video surveillance.
\newblock {\em IEEE transactions on cybernetics}, 51(10):4822--4833, 2020.

\bibitem{idrees2013multi}
Haroon Idrees, Imran Saleemi, Cody Seibert, and Mubarak Shah.
\newblock Multi-source multi-scale counting in extremely dense crowd images.
\newblock In {\em CVPR}, pages 2547--2554, 2013.

\bibitem{idrees2018composition}
Haroon Idrees, Muhmmad Tayyab, Kishan Athrey, Dong Zhang, Somaya Al-Maadeed,
  Nasir Rajpoot, and Mubarak Shah.
\newblock Composition loss for counting, density map estimation and
  localization in dense crowds.
\newblock In {\em ECCV}, pages 532--546, 2018.

\bibitem{jiang2018acquisition}
Borui Jiang, Ruixuan Luo, Jiayuan Mao, Tete Xiao, and Yuning Jiang.
\newblock Acquisition of localization confidence for accurate object detection.
\newblock In {\em ECCV}, pages 784--799, 2018.

\bibitem{kingma2014adam}
Diederik~P Kingma and Jimmy Ba.
\newblock Adam: A method for stochastic optimization.
\newblock {\em arXiv preprint arXiv:1412.6980}, 2014.

\bibitem{kratz2009anomaly}
Louis Kratz and Ko Nishino.
\newblock Anomaly detection in extremely crowded scenes using spatio-temporal
  motion pattern models.
\newblock In {\em CVPR}, pages 1446--1453. IEEE, 2009.

\bibitem{krizhevsky2012imagenet}
Alex Krizhevsky, Ilya Sutskever, and Geoffrey~E Hinton.
\newblock Imagenet classification with deep convolutional neural networks.
\newblock {\em NeurIPS}, 25:1097--1105, 2012.

\bibitem{li2021video}
Haopeng Li, Lingbo Liu, Kunlin Yang, Shinan Liu, Junyu Gao, Bin Zhao, Rui
  Zhang, and Jun Hou.
\newblock Video crowd localization with multi-focus gaussian neighbor attention
  and a large-scale benchmark.
\newblock {\em arXiv preprint arXiv:2107.08645}, 2021.

\bibitem{8648370}
Wei Li, Hongliang Li, Qingbo Wu, Xiaoyu Chen, and King~Ngi Ngan.
\newblock Simultaneously detecting and counting dense vehicles from drone
  images.
\newblock {\em IEEE TIE}, 66(12):9651--9662, 2019.

\bibitem{li2018csrnet}
Yuhong Li, Xiaofan Zhang, and Deming Chen.
\newblock Csrnet: Dilated convolutional neural networks for understanding the
  highly congested scenes.
\newblock In {\em CVPR}, pages 1091--1100, 2018.

\bibitem{8283783}
Ningxin Liang, Guile Wu, Wenxiong Kang, Zhiyong Wang, and David~Dagan Feng.
\newblock Real-time long-term tracking with prediction-detection-correction.
\newblock {\em IEEE Transactions on Multimedia}, 20(9):2289--2302, 2018.

\bibitem{6520940}
Cao Lijun and Huang Kaiqi.
\newblock Video-based crowd density estimation and prediction system for
  wide-area surveillance.
\newblock {\em China Communications}, 10(5):79--88, 2013.

\bibitem{lin2017feature}
Tsung-Yi Lin, Piotr Doll{\'a}r, Ross Girshick, Kaiming He, Bharath Hariharan,
  and Serge Belongie.
\newblock Feature pyramid networks for object detection.
\newblock In {\em CVPR}, pages 2117--2125, 2017.

\bibitem{liu2015bayesian}
Bo Liu and Nuno Vasconcelos.
\newblock Bayesian model adaptation for crowd counts.
\newblock In {\em ICCV}, pages 4175--4183, 2015.

\bibitem{liu2019recurrent}
Chenchen Liu, Xinyu Weng, and Yadong Mu.
\newblock Recurrent attentive zooming for joint crowd counting and precise
  localization.
\newblock In {\em CVPR}, pages 1217--1226, 2019.

\bibitem{liu2021cross}
Lingbo Liu, Jiaqi Chen, Hefeng Wu, Guanbin Li, Chenglong Li, and Liang Lin.
\newblock Cross-modal collaborative representation learning and a large-scale
  rgbt benchmark for crowd counting.
\newblock In {\em Proceedings of the IEEE/CVF Conference on Computer Vision and
  Pattern Recognition}, pages 4823--4833, 2021.

\bibitem{liu2019crowd}
Lingbo Liu, Zhilin Qiu, Guanbin Li, Shufan Liu, Wanli Ouyang, and Liang Lin.
\newblock Crowd counting with deep structured scale integration network.
\newblock In {\em Proceedings of the IEEE/CVF international conference on
  computer vision}, pages 1774--1783, 2019.

\bibitem{liu2020estimating}
Weizhe Liu, Mathieu Salzmann, and Pascal Fua.
\newblock Estimating people flows to better count them in crowded scenes.
\newblock In {\em ECCV}, pages 723--740. Springer, 2020.

\bibitem{liu2019point}
Yuting Liu, Miaojing Shi, Qijun Zhao, and Xiaofang Wang.
\newblock Point in, box out: Beyond counting persons in crowds.
\newblock In {\em CVPR}, pages 6469--6478, 2019.

\bibitem{luo2020multiple}
Wenhan Luo, Junliang Xing, Anton Milan, Xiaoqin Zhang, Wei Liu, and Tae-Kyun
  Kim.
\newblock Multiple object tracking: A literature review.
\newblock {\em Artificial Intelligence}, page 103448, 2020.

\bibitem{ma2015counting}
Zheng Ma and Antoni~B Chan.
\newblock Counting people crossing a line using integer programming and local
  features.
\newblock {\em IEEE TCSVT}, 26(10):1955--1969, 2015.

\bibitem{ma2019bayesian}
Zhiheng Ma, Xing Wei, Xiaopeng Hong, and Yihong Gong.
\newblock Bayesian loss for crowd count estimation with point supervision.
\newblock In {\em ICCV}, pages 6142--6151, 2019.

\bibitem{munkres1957algorithms}
James Munkres.
\newblock Algorithms for the assignment and transportation problems.
\newblock {\em Journal of the society for industrial and applied mathematics},
  5(1):32--38, 1957.

\bibitem{paszke2019pytorch}
Adam Paszke, Sam Gross, Francisco Massa, Adam Lerer, James Bradbury, Gregory
  Chanan, Trevor Killeen, Zeming Lin, Natalia Gimelshein, Luca Antiga, et~al.
\newblock Pytorch: An imperative style, high-performance deep learning library.
\newblock {\em NeurIPS}, 32:8026--8037, 2019.

\bibitem{peyre2019computational}
Gabriel Peyr{\'e}, Marco Cuturi, et~al.
\newblock Computational optimal transport: With applications to data science.
\newblock {\em Foundations and Trends{\textregistered} in Machine Learning},
  11(5-6):355--607, 2019.

\bibitem{ren2020tracking}
Weihong Ren, Xinchao Wang, Jiandong Tian, Yandong Tang, and Antoni~B Chan.
\newblock Tracking-by-counting: Using network flows on crowd density maps for
  tracking multiple targets.
\newblock {\em IEEE TIP}, 30:1439--1452, 2020.

\bibitem{sam2017switching}
Deepak~Babu Sam, Shiv Surya, and R~Venkatesh Babu.
\newblock Switching convolutional neural network for crowd counting.
\newblock In {\em CVPR}, pages 4031--4039, 2017.

\bibitem{sarlin2020superglue}
Paul-Edouard Sarlin, Daniel DeTone, Tomasz Malisiewicz, and Andrew Rabinovich.
\newblock Superglue: Learning feature matching with graph neural networks.
\newblock In {\em CVPR}, pages 4938--4947, 2020.

\bibitem{simonyan2014very}
Karen Simonyan and Andrew Zisserman.
\newblock Very deep convolutional networks for large-scale image recognition.
\newblock {\em arXiv preprint arXiv:1409.1556}, 2014.

\bibitem{sinkhorn1967concerning}
Richard Sinkhorn and Paul Knopp.
\newblock Concerning nonnegative matrices and doubly stochastic matrices.
\newblock {\em Pacific Journal of Mathematics}, 21(2):343--348, 1967.

\bibitem{song2021rethinking}
Qingyu Song, Changan Wang, Zhengkai Jiang, Yabiao Wang, Ying Tai, Chengjie
  Wang, Jilin Li, Feiyue Huang, and Yang Wu.
\newblock Rethinking counting and localization in crowds: A purely point-based
  framework.
\newblock In {\em ICCV}, pages 3365--3374, 2021.

\bibitem{sun2019benchmark}
Shijie Sun, Naveed Akhtar, Huansheng Song, Chaoyang Zhang, Jianxin Li, and
  Ajmal Mian.
\newblock Benchmark data and method for real-time people counting in cluttered
  scenes using depth sensors.
\newblock {\em IEEE TITS}, 20(10):3599--3612, 2019.

\bibitem{sundararaman2021tracking}
Ramana Sundararaman, Cedric De~Almeida~Braga, Eric Marchand, and Julien Pettre.
\newblock Tracking pedestrian heads in dense crowd.
\newblock In {\em CVPR}, pages 3865--3875, 2021.

\bibitem{un2019world}
UN.
\newblock World population prospects 2019, 2019.

\bibitem{phdtt}
Xuan-Thuy Vo.
\newblock Phdtt results.
\newblock [Online].
\newblock \url{https://motchallenge.net/method/HT=7&chl=21}.

\bibitem{wan2021generalized}
Jia Wan, Ziquan Liu, and Antoni~B Chan.
\newblock A generalized loss function for crowd counting and localization.
\newblock In {\em CVPR}, pages 1974--1983, 2021.

\bibitem{wan2019residual}
Jia Wan, Wenhan Luo, Baoyuan Wu, Antoni~B Chan, and Wei Liu.
\newblock Residual regression with semantic prior for crowd counting.
\newblock In {\em CVPR}, pages 4036--4045, 2019.

\bibitem{wang2020distribution}
Boyu Wang, Huidong Liu, Dimitris Samaras, and Minh Hoai.
\newblock Distribution matching for crowd counting.
\newblock {\em arXiv preprint arXiv:2009.13077}, 2020.

\bibitem{gao2020nwpu}
Qi Wang, Junyu Gao, Wei Lin, and Xuelong Li.
\newblock Nwpu-crowd: A large-scale benchmark for crowd counting and
  localization.
\newblock {\em IEEE TPAMI}, 2020.

\bibitem{9337191}
Qi Wang, Tao Han, Junyu Gao, and Yuan Yuan.
\newblock Neuron linear transformation: Modeling the domain shift for crowd
  counting.
\newblock {\em IEEE Transactions on Neural Networks and Learning Systems},
  2021.

\bibitem{wu2020fast}
Xingjiao Wu, Baohan Xu, Yingbin Zheng, Hao Ye, Jing Yang, and Liang He.
\newblock Fast video crowd counting with a temporal aware network.
\newblock {\em Neurocomputing}, 403:13--20, 2020.

\bibitem{xiong2017spatiotemporal}
Feng Xiong, Xingjian Shi, and Dit-Yan Yeung.
\newblock Spatiotemporal modeling for crowd counting in videos.
\newblock In {\em ICCV}, pages 5151--5159, 2017.

\bibitem{zhang2021cross}
Qi Zhang, Wei Lin, and Antoni~B Chan.
\newblock Cross-view cross-scene multi-view crowd counting.
\newblock In {\em CVPR}, pages 557--567, 2021.

\bibitem{zhang2021fairmot}
Yifu Zhang, Chunyu Wang, Xinggang Wang, Wenjun Zeng, and Wenyu Liu.
\newblock Fairmot: On the fairness of detection and re-identification in
  multiple object tracking.
\newblock {\em IJCV}, pages 1--19, 2021.

\bibitem{zhang2016single}
Yingying Zhang, Desen Zhou, Siqin Chen, Shenghua Gao, and Yi Ma.
\newblock Single-image crowd counting via multi-column convolutional neural
  network.
\newblock In {\em CVPR}, pages 589--597, 2016.

\bibitem{zhao2016crossing}
Zhuoyi Zhao, Hongsheng Li, Rui Zhao, and Xiaogang Wang.
\newblock Crossing-line crowd counting with two-phase deep neural networks.
\newblock In {\em ECCV}, pages 712--726. Springer, 2016.

\bibitem{zheng2018cross}
Huicheng Zheng, Zijian Lin, Jiepeng Cen, Zeyu Wu, and Yadan Zhao.
\newblock Cross-line pedestrian counting based on spatially-consistent
  two-stage local crowd density estimation and accumulation.
\newblock {\em IEEE TCSVT}, 29(3):787--799, 2018.

\end{thebibliography}
}
\appendix
\begin{onecolumn}

The supplementary file provides more details for paper `` DR.VIC: Decomposition and Reasoning for Video Individual Counting", including the following aspects.
\begin{enumerate}

\item [1)] \textbf{Extra Experiments},
\item [2)] \textbf{Detailed Network Architectures},
\item [3)] \textbf{Limitations},
\item [4)] \textbf{Potential Negative Societal  Impact},
\item [5)] \textbf{More Visualization Results}.
\item [6)] \textbf{Video Demonstration}
\end{enumerate}

\section{Extra Experiments} \label{}
\subsection{MIAE and MOAE}
In the experiment part, we propose the Mean Inflow/Outflow Absolute Error (MIAE/MOAE) to evaluate the inflow and outflow estimation performance in the pair-wise images. Their detailed formulations are:
\begin{eqnarray} \label{eq:MIAE}
MIAE \text{@} \uptau = \frac{\sum\limits_{i=1}^{K}  \sum\limits_{t=\uptau}^{T_i-1}|\hat{N}^{+}_i(t) - N^{+}_i(t) |}{\sum_{i=1}^{K}{T_i-\uptau}} , \quad
MOAE \text{@} \uptau = \frac{\sum\limits_{i=1}^{K}  \sum\limits_{t=\uptau}^{T_i-1}|\hat{N}^{-}_i(t) - N^{-}_i(t) |}{\sum_{i=1}^{K}{T_i-\uptau}
}\end{eqnarray}

where $N^{+}_i(t)$ and $\hat{N}^{+}_i(t)$ represent the ground truth and predicted pedestrian inflow of frame $\mathbf{I}_t$ compared with frame $\mathbf{I}_{t-\uptau}$, respectively. Similarly, $N^{-}_i(t)$ and $\hat{N}^{-}_i(t)$ are the ground truth  and prediction for pedestrian outflow. $T_i$ is the total number of frames of the $i$-th video. $\uptau$ is the time interval (\ie frame intervals) for sampling frames.

\subsection{Scenes Categorization of SenseCrowd}
To comprehensively evaluate the performance of our method on diverse scenes, we manually annotate
the 634 videos in SenseCrowd with different scene labels from four perspectives, including:
\begin{enumerate}[1)]
\item Location: videos are classified into six categories by their locations. \cref{fig:dataset} a) shows the proportional distribution.
\item Density: Pedestrian density is a vital factor influencing the counting performance. Videos are divided into five classes according to the crowd density as shown in \cref{fig:dataset} b).
\item Time: $76\%$ of videos are recorded in daylight and remaining $24\%$ of videos at night.
\item Space: $77\%$ of videos are captured in outdoor scenes and $23\%$ in indoor scenes.
\end{enumerate}

\begin{figure}[tbp]
\centering
\includegraphics[width=0.9\textwidth]{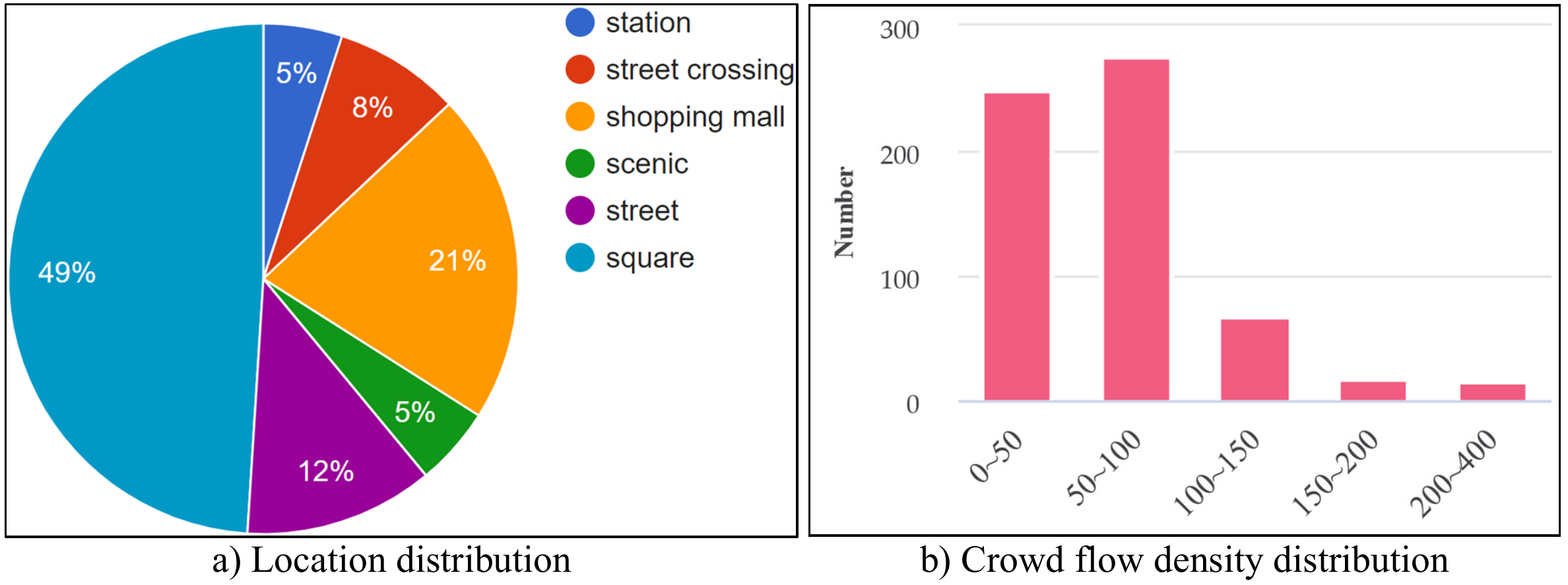}
\caption{ The pie chart of the captured video location and the statistical histogram of pedestrian counts on SenseCrowd dataset. }
\label{fig:dataset}
\end{figure}

\subsection{Performance of Diverse Scenes on SenseCrowd}
Based on the scene labels, we test the counting performance with different scenes and report the MAE in \cref{tab:SenseCrowd_other}. DRNet outperforms the tracking-based methods \cite{zhang2021fairmot,sundararaman2021tracking} and density-based methods (single image counting baselines and \cite{munkres1957algorithms}) in all scenes with obvious gaps. Take the Square videos as an example, DRNet achieves 8.4 MAE, which is a very low estimation error among all tested methods. At the same time, other methods, such as directly sampling the density map for video individual counting, lead to very poor results. 

\cref{tab:sup_CroHead} reports the nunmber of pedestrians on CroHD videos by directly counting the total identificaitons in the uploaded rtracking results \cite{MOTchallenge}.
We find that these figures are far more than the ground truth.     
Although we can get more reasonable results by searching an optimal sample rate when performing tracking, it still has a huge margin of error. Hence, the tracking-based methods is not robust and advisable when applying it to a long-time video for counting pedestrian number.

\begin{table*}[htbp]	
\centering
\setlength{\tabcolsep}{2.7mm}{\begin{tabular}{cIccccccIccIcc}
		\whline
		\multirow{2}{*}{Methods}  & \multicolumn{6}{cI}{Location } & \multicolumn{2}{cI}{Time} & \multicolumn{2}{c}{Space}\\
		\cline{2-11}
		&L0&L1&L2&L3&L4&L5&Day&Night& Indoor&Outdoor \\
		\whline
		FairMOT \cite{zhang2021fairmot}   &23.4&25.9&38.8&61.6&32.7&51.7 &27.3&35.6 &27.7&34.9\\
		HeadHunter-T \cite{sundararaman2021tracking}  &22.8&28.6&46.9&61.6&29.0&56.8 &29.2&32.8 &31.7&29.5\\
		
		\whline
		SICC (sampling)  &177.2&139.7&210.1&233.7&135.7&229.1 &178.8&152.5&143.0&182.1\\
		SICC (maximum)  &26.6&26.9&26.2&26.8&34.6&40.4 &29.7&24.1&27.0&28.9\\
		LOI  \cite{munkres1957algorithms}  &24.4&21.3&20.7&28.5&22.1&44.3 &26.8&17.8&22.6&25.4\\
		\whline
		DRNet &\textbf{8.4}&\textbf{11.2}&\textbf{18.1}&\textbf{33.6}&\textbf{11.2}&\textbf{29.9} &\textbf{11.8}&\textbf{14.1} &\textbf{12.6}&\textbf{12.2}\\
		\whline
\end{tabular}}
\caption{Video Individual Counting performance on SenseCrowd dataset measured by MAE. $L0\sim L5$ represents six location categories: square, shopping mall, street crosses, scenic, street, and station, respectively.}
\label{tab:SenseCrowd_other}
\end{table*}

\begin{table*}[htbp]	
\small
\centering
\setlength{\tabcolsep}{2.5mm}{\begin{tabular}{cIcccccIccc}
		\whline
		
		\multirow{2}{*}{Methods}& CroHD11 &CroHD12 &CroHD13&CroHD14&CroHD15 &\multirow{2}{*}{MAE$\downarrow$} &\multirow{2}{*}{MSE$\downarrow$}&\multirow{2}{*}{MRAE(\%)$\downarrow$} \\
		\cline{2-6}
		&   \underline{133} &\underline{737} &\underline{734}&\underline{1040}&\underline{321} & &&\\
		\whline
		FairMOT \cite{zhang2021fairmot} &366 &3215 &7011 &2626 &2337    &2518.0&3230.3&428.1 \\
		HeadHunter-T \cite{sundararaman2021tracking}  &307 &2145 &2556 &1531 &888 &892.4&1085.8&166.0  \\
		\whline
\end{tabular}}
\caption{Video individual counting performance of thracking-based methods. The \underline{underline} fonts represent the number of ground truth pedestrians. These results show that the existing tracking methods cannot be directly appplied to this new task.}
\label{tab:sup_CroHead}
\end{table*}

\section{Network Architectures}
\cref{Table:CP1} elaborates the detailed network architecture of DRNet, including the image representation backbone and head descriptor extraction module, which is consisted a head localization branch and a descriptor generation branch.

\textbf{Image Representation Backbone.}
\cref{Table:CP1} explains the VGG16 configurations in DRNet. In this table, ``k(3,3)-c256-s1-BN-R'' represents the convolutional or de-convolutional  operation with kernel size of $3 \times 3$, output channels of $256$, and stride size of $1$. The ``BN'' and ``R'' mean that the Batch Normalization and ReLU layer are added to this convolutional layer. With the VGG-16 \cite{simonyan2014very} backbone, we output three stages features and fuse them with the Feature Pyramid Networks (FPN) \cite{lin2017feature}.

\textbf{Head Localization Branch.}
\cref{Table:CP1} also explains the configurations for extracting head proposals. ``ResBlock-c256-s1-BN-R'' represents a residual CNN module with three convolution layers, $256$ output channels, and stride size of $1$. This branch finally outputs one-channel density map with the same shape as the input image.

\textbf{Descriptor Generation Branch.}
Similarly, we use two Residual modules and two convolution layers to further refine the feature map for head descriptor generation, which receives 384-channel feature maps and produces a 256-channel feature maps.

\begin{table}[htbp]
\centering

\begin{tabular}{cIc}
	\whline
	\multicolumn{2}{c}{\textbf{Image Representation Backbone} (VGG16)} \\
	\whline
	\multirow{3}{*}{Stage1 (1/4)} &conv1: [k(3,3)-c64-s1-BN-R]  \\
	&{...}\\
	&conv7: [k(3,3)-c256-s1-BN-R]\\
	\whline
	\multirow{3}{*}{Stage2 (1/8)}    &{...}\\
	
	&conv10: [k(3,3)-c512-s1-BN-R]\\
	\whline
	\multirow{3}{*}{Stage3 (1/16)}    &{...}\\
	
	&conv13: [k(3,3)-c512-s1-BN-R]\\
	\whline
	\multicolumn{2}{c}{\textbf{output channels: [Stage1:256, Stage2:512, Stage3:512]}}\\
	\whline
	FPN Module (output channels: 576) & FPN Module (output channels: 384)\\
	\whline
	\multicolumn{2}{c}{\textbf{Head Descriptor Extraction}}\\
	\whline
	\textbf{Head Localization Branch} & \textbf{Descriptor Generation Branch} \\
	\whline
	Dropout2d(0.2)                      & Dropout2d(0.2)\\
	ResBlock-c256-s1-BN-R                   & ResBlock-c384-s1-BN-R\\
	ResBlock-c128-s1-BN-R                   & ResBlock-c256-s1-BN-R  \\
	deconv:k(2,2)-c64-s2-BN-R           & conv:k(3,3)-c256-s1-BN-R\\
	conv:k(3,3)-c32-s1-BN-R             & conv:k(3,3)-c256-s1\\
	deconv:k(2,2)-c16-s2-BN-R\\
	conv:k(3,3)-c1-s1-R\\
	\whline
	
\end{tabular}
\caption{Detailed network architecture of DRNet.}
\label{Table:CP1}
\end{table}

\section{Limitations}
While our work achieves promising video individual counting, it has two limitations:

\textbf{Simplification:} In this work, we propose the direction to simplify the video individual counting to inflow estimation between image pairs according to the observation in real world datasets. 
While effective, this direction can not handle the cases that people pop in and out of the scene in a very shot time (such as 1 second) and the cases that people re-enter the scene after a relative long period (e.g., 1 minute). Thus, this direction is not capable of generating 100\% accurate counting. 

\textbf{Evaluation on very long videos:} In the real world applications, counting on very long videos (e.g., 1 hour) are useful. However, our experiments does not cover this case due to the limitations in datasets. We will collect and label long videos for this task in the future work.

\section{Potential negative societal impact}

\textbf{Employment impact:} The development of intelligent systems will inevitably require fewer human resources. That means fewer people will be hired in some related fields. For example, the video individual counting discussed in this paper is usually done manually by some professional staffs. Once this technology is applied in practice, some security personnel and management personnel may be affected as the job opportunity may be cut down. This technology can also be transferred to some traditional industries, such as commodity statistics, which will also affect the employment of some workers.

\textbf{Environmental impact:} The training of the model involved in this technology requires considerable electrical support, which would consume a certain amount of energy. We suggest using clean energy for decreasing the impact on environment.

\section{More visualization results}
\cref{fig:result1} and \cref{fig:result2} present more visualization samples on the SenseCrowd test set. Overall, these samples on a variety of scenes demonstrate that DRNet achieves promising inflow (and additional  outflow) reasoning performance, which ensures the success of DRNet in video individual counting. However, we can still observe that some difficult scenes (e.g., high density, occlusion, person multi-view and scale variations \etc) have a lot of rooms for improvement, which also points out a direction for future research.

\begin{figure}
\centering
\includegraphics[width=0.9\textwidth]{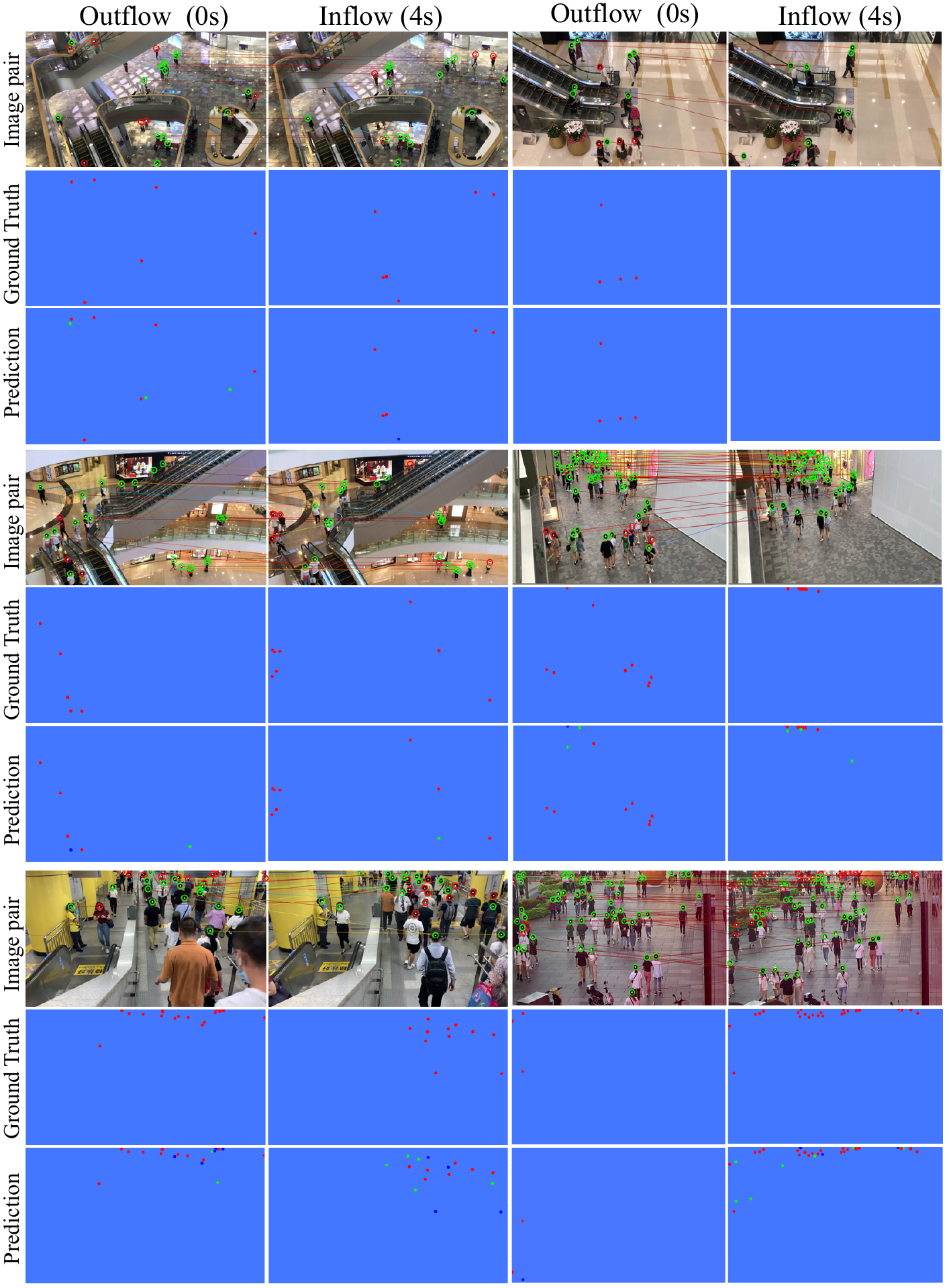}
\caption{ Visualization samples of SenseCrowd. The green and red circles in the image pairs denote matched and unmatched pedestrians, respectively.
	The {\color{red}{red}}, {\color{blue}{blue}}, and {\color{green}{green}} points in prediction results respectively denote correctly identified flows (inflow or outflow in the corresponding columns), missed flows, and over-counted flows, respectively.}
\label{fig:result1}
\end{figure}

\begin{figure}
\centering
\includegraphics[width=0.95\textwidth]{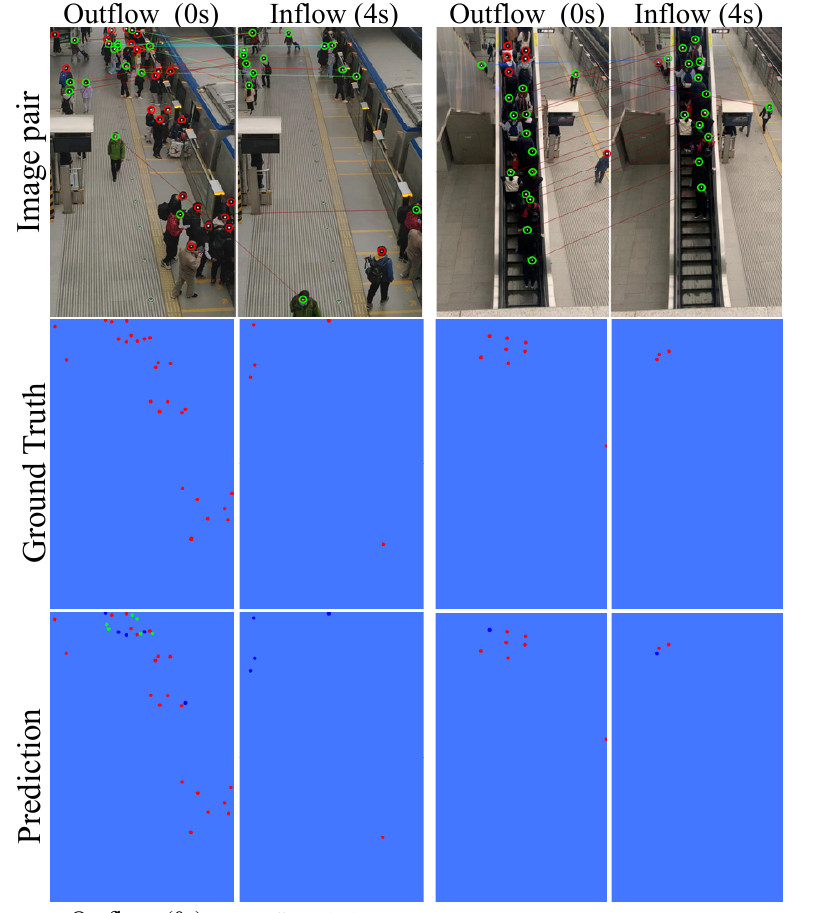}
\caption{ Visualization samples of SenseCrowd. The green and red circles in the image pairs denote matched and unmatched pedestrians, respectively.
	The {\color{red}{red}}, {\color{blue}{blue}}, and {\color{green}{green}} points in prediction results respectively denote correctly identified flows (inflow or outflow in the corresponding columns), missed flows, and over-counted flows, respectively.}
\label{fig:result2}
\end{figure}

\section{Video demonstration}
We also make a video demo to showcase the performance of DRNet for Video Individual Counting, please check it if interested.
Fig. \ref{Fig:demo} shows a screenshot of the demo, which demonstrates the pedestrian count performance with the time goes by. In the video clip, it can be found that DRNet can effectively count people for several minutes even in the poor light scene. Besides, we also notice that the accumulated  error will make the prediction number go away from the ground truth number gradually, which also tell us the future concentration in this task is how to further eliminate the accumulated error.
Overall, this demonstration shows the video individual counting technique would be possible to help improving the social management in the near future. For watching the complete video, please move to \url{https://youtu.be/CIqexlvYT4g} or  \url{https://www.bilibili.com/video/BV1cY411H7hr/}.

\begin{figure}[h]
\centering
\includegraphics[scale=0.98]{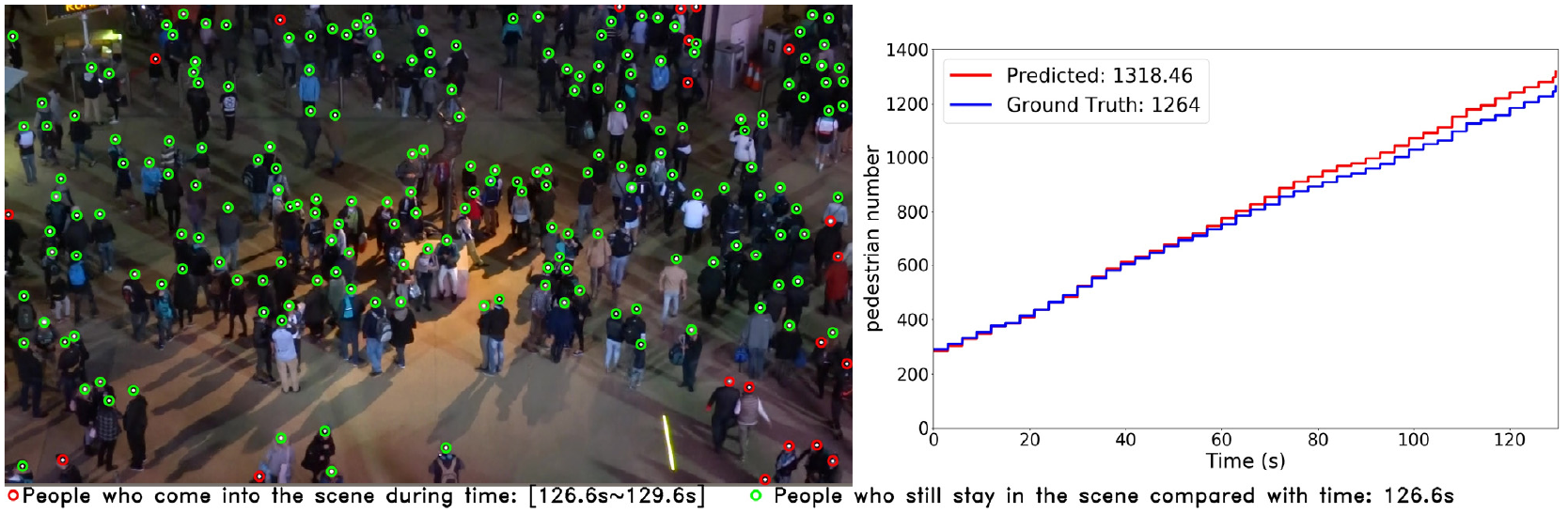}
\caption{The screenshots of the video demonstration. Left: {\color{red}{red}} circles depict the people who entered the scene from 3 seconds ago to now. {\color{green}{green}} circles represent the people who still stay in the scene compared with the frame at 3 seconds ago. Right:  {\color{red}{red}} curve and {\color{blue}{blue}} curve represent the predicted and ground truth accumulated pedestrian count (including the initial crowd count and the later inflow-crowd) from time 0, respectively. } 
\label{Fig:demo}
\end{figure}

\end{onecolumn}
\end{document}